\documentclass{sn-jnl}

\usepackage{booktabs,multirow,array,caption}
\usepackage{tabularx}
\usepackage{graphicx}
\usepackage{float}
\usepackage{amsmath}
\usepackage{amssymb}
\usepackage{amsthm}
\usepackage{xcolor}
\usepackage{bigints}

\newcommand\eatpunct[1]{}
\usepackage{dsfont}
\usepackage{subcaption}
\usepackage[algo2e]{algorithm2e}
\RestyleAlgo{ruled} 
\usepackage{enumerate}
\usepackage{caption}

\newtheorem{proposition}{Proposition}[section]

\newtheorem{theorem}{Theorem}[section]

\allowdisplaybreaks

\jyear{2022}
\begin{document}

\title[Plant Species Recognition with Optimized 3DPNN and VOTCSW]{Plant Species Recognition with Optimized 3D Polynomial Neural Networks and Variably Overlapping Time--Coherent Sliding Window}


\author*[1]{\fnm{Habib} \sur{Ben Abdallah}}\email{benabdallah-h@webmail.uwinnipeg.ca}

\author[1]{\fnm{Christopher J.} \sur{Henry}}\email{ch.henry@uwinnipeg.ca}

\author[1]{\fnm{Sheela} \sur{Ramanna}}\email{s.ramanna@uwinnipeg.ca}

\affil[1]{\orgdiv{Department of Applied Computer Science}, \orgname{University of Winnipeg}, \orgaddress{\street{515 Portage Ave}, \city{Winnipeg}, \postcode{R3B 2E9}, \state{Manitoba}, \country{Canada}}}


\abstract{Plant species recognition is a primordial task that forms the basis of solving several plant-related computer-vision problems such as disease detection or growth monitoring. However, a lack of voluminous datasets that cover specific needs is observed in the digital agriculture community. Therefore, the EAGL—I system was developed to rapidly create massive labeled datasets of plants intended to be commonly used by farmers and researchers to create AI-driven solutions in agriculture. As a result, a publicly available plant species recognition dataset composed of 40,000 images with different sizes consisting of 8 plant species was created with the system in order to demonstrate its capabilities. This paper proposes a novel method, called Variably Overlapping Time—Coherent Sliding Window (VOTCSW), that transforms a dataset composed of images with variable size to a 3D representation with a common fixed size that is suitable for convolutional neural networks, and demonstrates that this representation is more informative than resizing the images of the dataset to a given size. We theoretically formalized the use cases of the method as well as its inherent properties and we proved that it has an oversampling and a regularization effect on the data. By combining the VOTCSW method with the 3D extension of a recently proposed machine learning model called 1-Dimensional Polynomial Neural Networks, we were able to create a model that achieved a state-of-the-art accuracy of 99.9\% on the dataset created by the EAGL—I system, surpassing well-known architectures such as ResNet and Inception. In addition, we created a heuristic algorithm that enables the degree reduction of any pre-trained N-Dimensional Polynomial Neural Network and which compresses it without altering its performance, thus making the model faster and lighter. Furthermore, we established that the currently available dataset could not be used for machine learning in its present form, due to a substantial class imbalance between the training set and the test set. Hence, we created a specific preprocessing and a model development framework that enabled us to improve the accuracy from 49.23\% to 99.9\%.}



\maketitle
\vspace{-20pt}
\section{Introduction}\label{chapter=plants:section=introduction}
\begin{figure}[!htpb]
\centering
\includegraphics[scale=0.089]{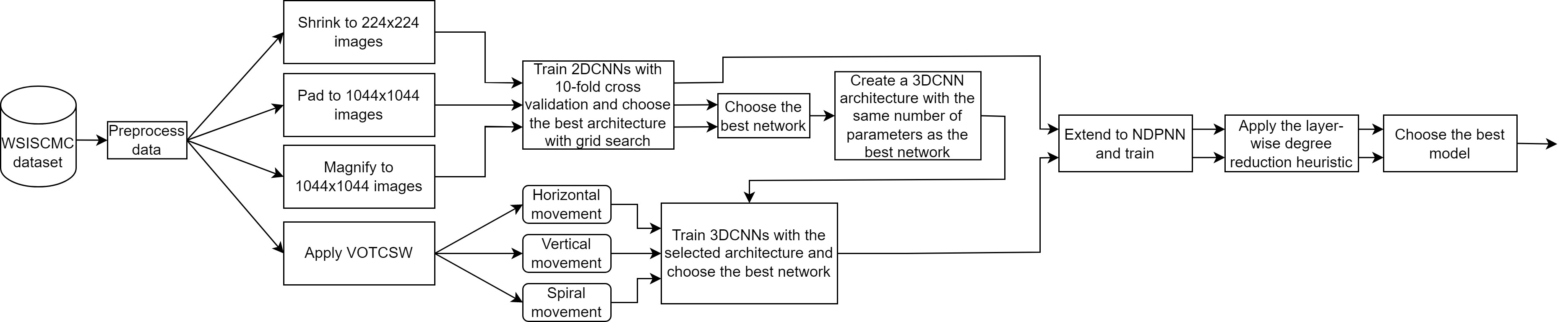}
\caption{Block diagram for producing a reliable plant species classification model.}
\label{fig:plants_framework}
\end{figure}

Agriculture is a sector that requires multi-disciplinary knowledge to steadily evolve \cite{intro:multi_disciplinary, intro:multi_disciplinary_1, intro:multi_disciplinary_2} since large-scale food production necessitates a deep understanding of every relevant plant species \cite{intro:deep_knowledge, intro:deep_knowledge_1, intro:deep_knowledge_2} and highly advanced machinery \cite{intro:machinery, intro:machinery_1, intro:machinery_2} to ensure an optimized yield. With the emergence and the recent practical successes of the Internet of Things, robotics and artificial intelligence \cite{intro:emergence}, the sector has observed a surge in smart farming solutions which has led the march to the fourth agricultural revolution \cite{intro:agriculture_4}. Consequently, new fields such as digital agriculture \cite{intro:digital_agriculture} and precision agriculture \cite{intro:precision_agriculture} have become intensively researched which has led to an ever-increasing pace in innovation. Although the integration of artificial intelligence is making its way to digital agricultural applications such as crop planting \cite{intro:planting, intro:planting_1, intro:planting_2} and harvesting \cite{intro:harvesting, intro:harvesting_1, intro:harvesting_2}, it is still limited to mechanical tasks that mainly require environmental awareness \cite{intro:machinery}. A number of issues that are highly concerning for farmers such as crop disease detection, plant-growth monitoring and need-based irrigation have only recently started to gain an interest in the machine learning community \cite{intro:machine_learning}. Despite the efforts to create automated systems to resolve such issues \cite{intro:efforts, intro:efforts_1}, only highly specialized systems are created which only work with specific species or variants under constrained conditions \cite{intro:disease, intro:growth, intro:irrigation}. Furthermore, due to the lack of voluminous labeled data for each species and for each specific issue \cite{intro:machine_learning}, automated systems mainly rely on advanced visual feature engineering that requires a team of plant specialists, and only partially rely on machine learning for using these features to extract useful information \cite{intro:feature_engineering, intro:feature_engineering_1, intro:feature_engineering_2}. In fact, the sheer diversity of plant species, variants, diseases, growth stages and growing conditions makes manual feature engineering unfeasible to cover every individual farmer's need. While some attempts at creating massive plants datasets have been successful \cite{intro:dataset, intro:dataset_1, intro:dataset_2}, their usefulness was limited by the species present in them and the initial task they were created for. For instance, a dataset containing species grown in a tropical biome for the purpose of species recognition has limited to no usefulness for farmers working in a grassland biome who want to establish the presence of a disease in certain plants.  
\newline\indent
As a result, the EAGL--I system \cite{intro:EAGL-I} has been recently proposed to automatically generate a high number of labeled images in a short time (1 image per second) in an effort to circumvent the problem of the lack of data for specific needs. Consequently, two massive datasets were created with this system \cite{intro:EAGL-I_big_dataset}; one that contains 1.2 million images of indoor-grown crops and weeds common to Canadian prairies, and one that contains 540,000 images of plants imaged in farmland. Also, a publicly available dataset called ``Weed seedling images of species common to Manitoba, Canada" (WSISCMC) \cite{intro:WSISCMC} which contains 40,000 images of 8 species that are very rarely represented in plants datasets was created with the EAGL--I system. The purpose of the creation of the dataset was to demonstrate the capability of the EAGL--I system to rapidly generate large amounts of data that are suitable for machine learning applications and that can be used to solve specific digital agriculture problems, such as the ability to recognize several species of grasses which are responsible for the loss of hundreds of millions of dollars. However, in \cite{intro:EAGL-I}, the validity of the dataset was only tested for a binary classification problem consisting in differentiating between grasses and non-grasses without determining the species of the given plant itself. While the model that solves this problem gives an indication on how well the samples of the dataset are distributed to allow for grass differentiation, it does not provide enough information on how detailed the samples of the dataset are at a granular level to allow the identification of the species they belong to. Indeed, the higher the number of mutually exclusive classes, the more distinctive and detailed the spatial features of the plants need to be. Therefore, we chose to tackle the problem of plant species recognition on the WSISCMC dataset leading to a solution that can help the early eradication of invasive species that are harmful to the growth of certain crops. Furthermore, this constitutes a step further in validating the dataset for machine learning applications, and by extension, the EAGL--I system for the creation of meaningful datasets. Thus, this work aims at maximizing the plant recognition accuracy on the WSISCMC dataset by creating a highly reliable and accurate network using the model development framework shown in Figure \ref{fig:plants_framework} in order to provide insight on how to improve the data acquisition process to produce cleaner samples for future massive datasets. Moreover, the particularity of the WSISCMC dataset lies in the fact that all images are captured in a blue background. This makes the dataset very versatile for domain translation techniques \cite{intro:translation} which can be used to artificially place the plants in an outdoor background representing a given agricultural field. As a result, despite the plants being captured in a lab-controlled environment, their usability can be extended to the real world. Thus, the species recognition knowledge acquired by a model trained on the WSISCMC can also be transferred to the real world via domain translation.
\newline\indent
To solve this problem, we decided to use a novel deep learning model that we recently developed called 1-Dimensional Polynomial Neural Network (1DPNN) \cite{intro:1DPNN} which was proven to encapsulate more information from its training set and to generalize better than a 1-Dimensional Convolutional Neural Network (1DCNN) in less time and with less memory than a 1DCNN. A major limitation of the 1DPNN model is that it can only process 1D signals and that there was no way to determine the polynomial degree of each of its layers upon solving a given problem. Therefore, we propose an extension of this model to 2D and 3D signals, such as images and videos respectively, and we generalize its denomination to NDPNN where ND stands for 1D, 2D or 3D. We also develop a heuristic algorithm that makes use of a polynomial degree reduction formula \cite{intro:reduction} that we have recently developed and that allows to determine the smallest degree of each layer of a pre-trained NDPNN that preserves its performance on its test set, thus, enabling it to use less memory and less computational power while maintaining the same performance on its test set. Moreover, since the WSISCMC dataset is composed of images with different sizes, we create a method that we call Variably Overlapping Time--Coherent Sliding Window (VOTCSW) which allows the transformation of images with variable sizes to a 3D representation with a common fixed size that is suitable for 3DCNNs and 3DPNNs or any machine learning model that processes 3D signals. The novelty of the VOTCSW lies in its ability to partially or even completely remove the issue of resizing images with different sizes in any given dataset to a common size, which always certainly causes a loss of information since the resolution is usually altered. In fact, most machine learning models require a predetermined fixed size for their inputs. Hence, the method is a valuable contribution to the fields of image processing and machine learning since it creates a common fixed-size 3D representation of images that have different sizes with minimal or even no loss of information. We also redistribute the samples of the dataset to maximize the learning efficiency of any machine learning model that may use it. Furthermore, we train various well-know architectures such as ResNetV2 \cite{intro:resnet}, InceptionV3 \cite{intro:inception} and Xception \cite{intro:xception}, and we evaluate the gain of using the VOTCSW method and the NDPNN degree reduction heuristic with respect to regular 2DCNNs architectures. 
\newline\indent
The contributions of this work are i) the extension of 1DPNNs to NDPNNs which can now be used on 2D and 3D signals such as images and videos, ii) the development of a heuristic algorithm for the degree reduction of pre-trained NDPNNs which creates lighter and faster NDPNNs with little to no compromise to their initial performances, iii) the formalization of the VOTCSW method which circumvents the need to resize images in a dataset that have different sizes by transforming each image to a 3D representation of a common fixed size that is suitable for 3DCNNs and 3DPNNs with minimal loss of information compared to regular resizing techniques and which improves their inference on the WSISCMC dataset, iv) the resampling of the WSISCMC dataset with respect to class distribution and size distribution in order to enhance the performance of any machine learning model trained on it, v) the creation of a NDPNN model development framework that makes use of the degree reduction heuristic and the VOTCSW method and allows the creation of the best fitting neural network architecture on the WSISCMC dataset, vi) the creation of a simple 3DPNN architecture that achieves a state-of-the-art 99.9\% accuracy on the WSISCMC dataset, and which outperforms highly complex neural network architectures such as ResNet50V2 and InceptionV3, in less time and with substantially less parameters, vii) and the determination of aberrant samples in the WSISCMC dataset which are not suitable for the single-plant species recognition task that this dataset was created for.
\newline\indent
The outline of this paper is as follows. Section \ref{chapter=plants:section=related work} discusses the most recent works in plant species recognition while Section \ref{chapter=plants:section=theoretical framework} introduces 1DPNNs and their extension to NDPNNs, develops the layer-wise degree reduction heuristic and formulates the theoretical foundation of the VOTCSW method. Section \ref{chapter=plants:section=experiments} presents the model development framework established to produce a highly accurate model for the WSISCMC dataset and discusses the results obtained while providing insights on how the methods developed in Section \ref{chapter=plants:section=theoretical framework} influenced them. Finally, Section \ref{chapter=plants:section=conclusion} presents a summary of what was achieved in this work, discusses the limitations of the methods and the models used, and proposes solutions to overcome these limitations.
\section{Related Work}\label{chapter=plants:section=related work}
\subsection{Plant species recognition}\label{chapter=plants:section=related work:sub=plants}
Plant species recognition is one of the most important tasks in the application of machine learning to digital agriculture. In this context, behaviour specific to a species will inform the identification of a plant's disease or the plant's need for resources such as water or light. The most common approach to identify the species of a given plant is to analyze its leaves \cite{related:leaves}. In fact, many public datasets \cite{intro:dataset, related:dataset} are only composed of leaves scanned in a uniform background. The methods that are mostly used rely either on combining feature engineering and machine learning classifiers or using deep learning models for both feature extraction and classification. 
\newline\indent
Indeed, Purohit et al. \cite{related:feature_engineering} created morphological features based on the geometric shape of any given leaf and used these features to discriminate between 33 species of plants using different classifiers. They achieved a state-of-the-art $95.42\%$ accuracy on the Flavia dataset \cite{intro:dataset} and they demonstrated that their morphological features are superior to color features or texture features. However, they did not compare the efficiency of their features to ones that are fully determined using deep learning models. On the contrary, Wang et al. \cite{related:leaf_deep_learning} created a novel multiscale convolutional neural network (CNN) with an attention mechanism that can filter out the global background information of the given leaf image and highlight its saliency which allows it to consistently outperform classification models based on morphological feature extraction, as well as classification models based on well-known CNN architectures. They explain that CNNs are better at extracting high-level and low-level features without the need to perform image preprocessing and that their model which combines both these types of features in an attempt to discover estimatable relationships outperforms regular CNNs. However, the models that were developed were only trained on the ICL dataset \cite{related:dataset} which only contains leaf images, and can therefore be hard to use on images of entire plants. 
\newline\indent
Mehdipour Ghazi et al. \cite{related:cnn_full_plant} have considered training versatile architectures of CNNs on the LifeCLEF 2015 \cite{intro:dataset_2} which contains $100,000$ images of plant organs from $1000$ species that are mostly captured outdoors. They proposed a data augmentation approach that randomly extracts and scales a number of random square patches from any given image before applying a rotation. These patches and the original image are then resized to a fixed size and the mean image is substracted from them in order to keep the most relevant features. The resulting images are then fed to a CNN model which outputs a prediction for each image. The prediction of the original image is then determined by summing these predictions together. This aggregation as well as fine-tuning pre-trained networks such as VGGNet \cite{related:vgg_net}, GoogLeNet \cite{related:google_net}, and AlexNet \cite{related:alex_net} with different hyperparameters controlling the number of weight updates, the batch size and the number of patches, allowed them to achieve state-of-the-art results on the dataset, notably by fusing VGGNet and GoogLeNet. The authors observe that, when training networks from scratch, simpler architectures are preferred to the kind of architectures they used in their work. Indeed, they could not train any network from scratch to produce satisfactory results, and they did not attempt to create simpler and more specialized architectures for the problem.
\newline\indent
The work presented in this paper differs from the related papers on plant species recognition in four ways:
\begin{enumerate}
\item The WSISCMC dataset contains images of entire plants that trace different growth stages such that a trained classifier may be able to integrate the temporal evolution of a species organs in its inference, and may produce features richer than the ones that are only determined from separate organs such as leaves. Furthermore, the images are all captured in a blue background which enables the transfer of the acquired knowledge by trained models to the real world via background subtraction or domain translation.
\item Instead of relying on resizing the images in the dataset, we theoretically develop and use the VOTCSW method which creates a 3D representation of a common fixed size for all the images in the dataset with little to no loss of information (compared to resizing).
\item We use a novel deep learning model called 1DPNN and we extend it to NDPNN.
\item Concomitantly, the neural network architectures that we used are constructed in various stages of simplicity to allow an ablation study. Starting from a simple 2DCNN architecture, a simple 3DCNN architecture relying on a different data representation is built, then both architectures are extended to NDPNNs. Finally, highly complex architectures such as InceptionV3 and ResNet50V2 are also considered.
\end{enumerate}
\subsection{Polynomial neural networks}\label{chapter=plants:section=related work:sub=poly}
The use of polynomials in neural networks has been motivated by their ease of use and their great expressiveness as well as the vast literature surrounding the use and definition of polynomials. Moreover, polynomials allow the definition of non-linear functions which are at the core of neural networks. In fact, neural networks with linear activation functions are reduced to a simple linear relationship between the input and the output, which is usually not enough for solving complex tasks that are non-linear in nature. As a result, researchers explored the use of polynomials in neural networks in different ways.
\newline\indent
With the simple use of second order and third order polynomials, the performance of neural networks for certain tasks drastically improved. Indeed, Wang et al. \cite{related:poly_degree_2} introduced the second-order response transform for visual recognition tasks which consists in combining the outputs of two network branches by summing them and adding a second order term that is a function of their element-wise product. They also altered the way residual blocks \cite{intro:resnet} work by introducing a term consisting of the square-root of the element-wise product between the block's input and output. Moreover, they applied these alterations on well-known neural network architectures such as LeNet \cite{related:le_net}, VGGNet \cite{related:vgg_net}, ResNet \cite{intro:resnet} and WRNet \cite{related:wr_net}, and trained them to perform various image recognition tasks, namely object recognition on CIFAR10 and CIFAR100 datasets \cite{related:cifar} and digit recognition on the SVHN dataset \cite{related:svhn}. In addition, they compared the prediction accuracy of the altered variants of the networks to their unaltered counterparts and they observed a consistent improvement in all the altered networks on all tasks with an extra computation overhead of less than 5\%. This proved that the second order response was an effective approach to widen and improve the representation capacity of neural networks.  Similarly, Hughes et al. \cite{related:poly_degree_2_rnn} introduced the use of second order polynomials in recurrent neural networks (RNNs) and outperformed state-of-the-art voice activity detection models. Moreover, Babiloni et al. \cite{related:poly_degree_3} showed that the use of third order polynomials could improve upon state-of-the-art models on image recognition, instance segmentation, and face detection with less computational overhead. However, these improvements were not verified for higher order polynomials. Nevertheless, quadratic deep networks were proved to be universal approximators \cite{related:universal} which paved the way for the exploration of the use of higher order polynomials in neural networks.
\newline\indent
Chrysos et al. \cite{related:deep_poly} introduced a new class of deep neural networks called deep polynomial neural networks such that the output of that network is a high-order polynomial of the input. For a high-dimensional input and a high-dimensional output, the polynomial is built such that each element of the output is a full polynomial expansion of a given order of all the elements of the input where the weights of that expansion are computed during training. This introduces a complexity in the number of parameters which grows exponentially with the order of the polynomial. Therefore, they perform 3 different tensor decompositions of the weights by using a recurrence relationship to lower the complexity and they also construct the polynomial as the composition of lower-degree polynomials such that each lower-degree polynomial is the input to the next lower-degree polynomial. By doing so, they are able to use different decompositions for the lower-degree polynomials which requires much less parameters for achieving the same order of approximation. The authors achieved state-of-the-art results on image generation,  audio classification, image classification and 3D mesh autoencoding. Nevertheless, deep polynomial neural networks do not support weight sharing which can help drastically reduce their complexity.
\newline\indent
Ben Abdallah et al. \cite{intro:1DPNN} proposed the 1DPNN model which is an extension of the 1DCNN model such that the output of a given neuron is computed by performing a convolution between weights associated to a given degree and the input of that neuron exponentiated to that same degree. The authors formalized the backpropagation equations of the 1DPNN model to allow these weights to be learned during training. Moreover, they evaluated the influence of various activation functions on the 1DPNN and analysed its behaviour with respect to the 1DCNN model on spoken digit recognition, musical note recognition and audio signal denoising. As a result, they were able to empirically prove that the 1DPNN model could learn more complex representations and achieve a better expressiveness with less computational power and less parameters than the 1DCNN model. They also determined that the use of polynomials introduced stability issues which could be solved with bounded activation functions. However, the 1DPNN model could not process 2D signals such as images and 3D signals such as videos. Furthermore, there was no known way to determine the optimal degree of each layer in a 1DPNN model other than experimentation and intuition.
\newline\indent
The work presented in this paper differs from the related papers on polynomial neural networks in two ways:
\begin{enumerate}
\item The use of polynomial neural networks with weight sharing on images and videos via 2DPNN and 3DPNN, respectively.
\item The use of a polynomial degree reduction formula \cite{intro:reduction} for the development of a heuristic that allows the optimization of the degree of the polynomials used in the approximation inherent to polynomial neural networks, hence, creating highly accurate lightweight and fast polynomial neural networks.
\end{enumerate}
\section{Theoretical Framework}\label{chapter=plants:section=theoretical framework}
In order to obtain the most accurate results on the plant species classification problem, we use a novel model called 1DPNN which we proposed for audio signal related problems. In this work, we generalize our model to be applied on 2D and 3D signals such as images and videos, and we show that the equations governing its inference and its learning do not change with the dimension of the signal. Hence, we refer to the generalized model as N-Dimensional PNN (NDPNN) since it can theoretically be used to process signals with any dimension. We also develop a heuristic algorithm for reducing the polynomial degree of each layer of a pre-trained NDPNN in a way that preserves its performance on a given test set. We also create a new method that we call variably overlapping time--coherent sliding window (VOTCSW) for changing the representation of images from a 2-dimensional grid of pixels to a 3-dimensional grid of pixels moving through time in order to solve the problem of the size variability of the images in the WSISCMC dataset considered in this problem. This section discusses the inference, the learning and the degree reduction of NDPNNs and details the foundation of the VOTCSW method.
\subsection{N-Dimensional Polynomial Neural Networks and Polynomial Degree Reduction}\label{chapter=plants:sub=NDPNN}
\begin{figure}[!htpb]
\centering
\includegraphics[scale=0.1]{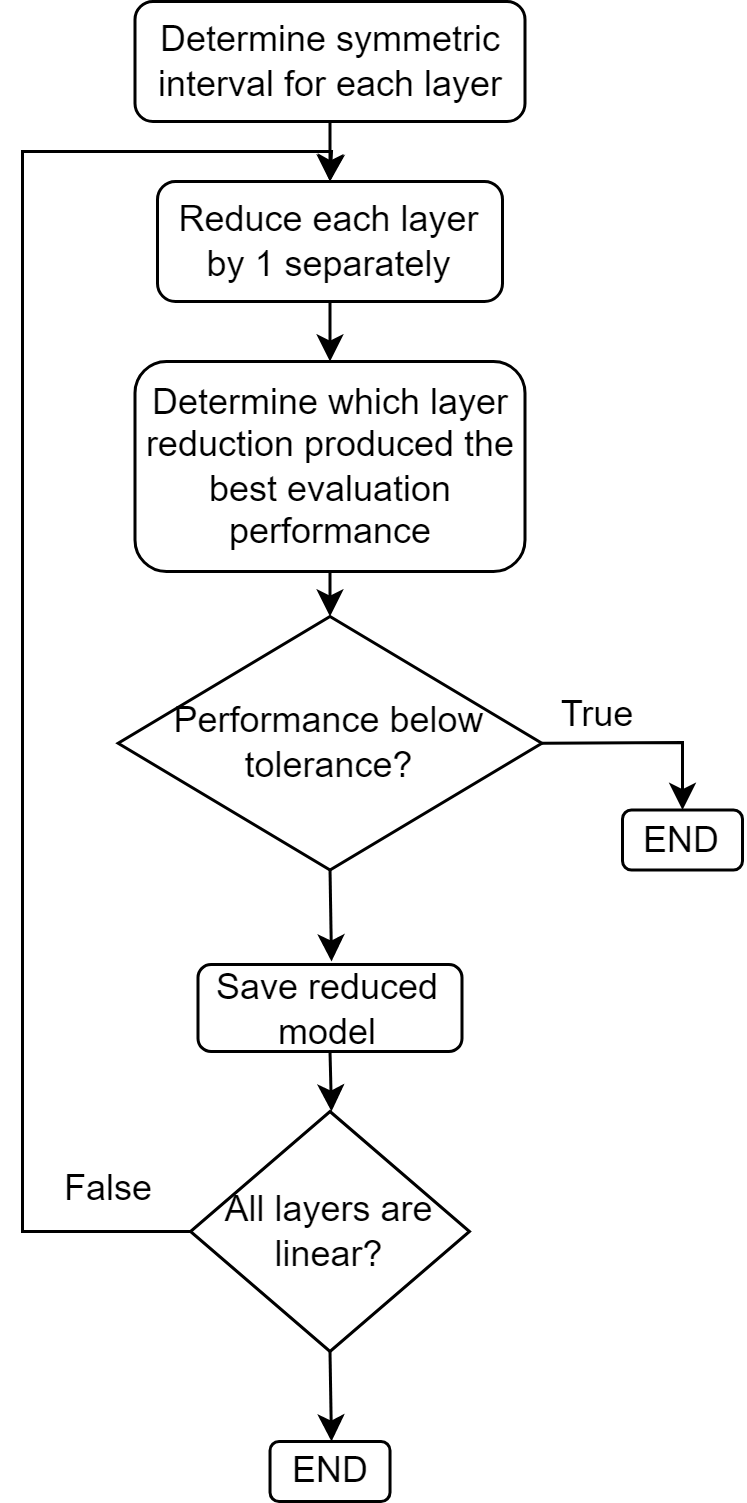}
\caption{Overview of the layer-wise degree reduction algorithm.}
\label{fig:degree_reduction_algorithm}
\end{figure}

1DPNNs are an extension of 1DCNNs such that each neuron creates a polynomial approximation of a kernel that is applied to its input. Given a network with $L$ layers, such that a layer $l\in[\![1,L]\!]$ contains $N_l$ neurons, a neuron $i\in[\![1,N_l]\!]$ in a layer $l$  produces an output $y_i^{(l)}$ from its previous layer's output $Y_{l-1}$. The neuron possess a bias $b_i^{(l)}$, an activation function $f_i^{(l)}$, and $D_l$ weight vectors $W_{id}^{(l)}$ corresponding to every exponentiation $d$ of its previous layer's output up to a degree $D_l$. Eq. (\ref{Forward propagation equation}) below shows the output of a 1DPNN neuron.
\begin{flalign}\label{Forward propagation equation}
\forall l\in[\![1,L]\!],\forall i\in[\![1,N_{l}]\!],\ 
y_{i}^{(l)}=f_{i}^{(l)}\left(\sum_{d=1}^{D_{l}}W_{id}^{(l)}*Y_{l-1}^{d}+b_{i}^{(l)}\right)=f_{i}^{(l)}\left(x_i^{(l)}\right),&&
\end{flalign}
where $*$ is the convolution operator, $Y_{l-1}^{d}=\underbrace{Y_{l-1}\odot\cdots\odot Y_{l-1}}_{d\ times}$, and $\odot$ is the Hadamard product \cite{intro:1DPNN}.
\newline\indent
The equation above shows that the weight $W_{id}^{(l)}$ corresponding to $Y_{l-1}^d$ can be of any dimension as long as the convolution with $Y_{l-1}^d$ remains valid. For instance, if $Y_{l-1}^d$ is a list of 2D feature maps, $W_{id}^{(l)}$ has to be a list of 2D filter masks. Therefore, the equation of a 1DPNN neuron can be applied on images and videos as long as the dimension of the weights are adjusted accordingly. Hence, the same equation governs the behavior of a 2DPNN, a 3DPNN and an NDPNN in general. Furthermore, this also applies to the gradient estimation of an NDPNN, so even the backpropagation remains unchanged. Moreover, we propose a method to reduce the degree of each layer after an NDPNN network is fully trained on a given dataset. This makes use of a fast polynomial degree reduction formula that we recently proposed \cite{intro:reduction} which can generate a polynomial of low degree that behaves the same as a given polynomial of higher degree on a symmetric interval. The method that we propose is a post-processing method that can compress a fully trained NDPNN, thus making it faster and lighter, without sacrificing its performance on the dataset it was trained on. The memory and computational efficiency gain mainly depend on the topology of the NDPNN and the performance loss tolerance. Although the polynomial degree reduction is performed on a symmetric interval and an NDPNN can use unbounded activation functions in general, we use the fact that, after training, the NDPNN weights do not change. So the input of each layer will be bounded in a certain interval when the NDPNN is fed with samples from the training set. The bounding interval may not be symmetric, but every interval is contained within a symmetric one, which allows us to properly use the polynomial degree reduction formula. Eq. (\ref{Forward propagation equation}) does not clearly show how the polynomial degree reduction can be achieved, which is why we consider the case of 1DPNNs to demonstrate the principle.
\newline\indent
In the 1D case, the output of a given layer $l$ is $Y_l$ which is a list containing the outputs of every neuron in that layer. The output of a neuron $i$ in that layer, $y_i^{(l)}=f_i^{(l)}\left(x_i^{(l)}\right)$, is a vector of size $M_l$. Similarly, the weights of that neuron with respect to the exponentiation degree $d$, $W_{id}^{(l)}$, is a list of $N_{l-1}$ vectors of size $K_l$. Therefore, expanding $x_i^{(l)}$ from Eq. (\ref{Forward propagation equation}) to compute each of its vector element independently produces:
\begin{flalign*}
\begin{split}
\forall m\in[\![0,M_l-1]\!],x_i^{(l)}(m)&=\sum_{d=1}^{D_l}\sum_{j=1}^{N_{l-1}}\sum_{k=0}^{K_l-1}w_{ijd}^{(l)}(k)\left(y_j^{(l-1)}(m+k)\right)^d+b_{i}^{(l)}\\
&=\sum_{j=1}^{N_{l-1}}\sum_{k=0}^{K_l-1}\left(\sum_{d=1}^{D_l}w_{ijd}^{(l)}(k)\left(y_j^{(l-1)}(m+k)\right)^d+\dfrac{b_i^{(l)}}{N_{l-1}K_l}\right)\\
&=\sum_{j=1}^{N_{l-1}}\sum_{k=0}^{K_l-1}P_{ijk}^{(l)}\left(y_j^{(l-1)}(m+k)\right),\\
\end{split}&&
\end{flalign*}
such that 
\begin{flalign*}
P_{ijk}^{(l)}(X)=\sum_{d=1}^{D_l}w_{ijd}^{(l)}(k)X^d+\dfrac{b_i^{(l)}}{N_{l-1}K_l},&&
\end{flalign*}
and $X$ is an indeterminate. This shows that the output of a neuron $i$ in a layer $l$ is the result of the summation of $N_{l-1}K_l$ distinct polynomials, and that the layer consists of $N_lN_{l-1}K_l$ distinct polynomials. In the general case of an NDPNN, the number of polynomials in a layer $l$ would be $N_lN_{l-1}$ multiplied by the receptive field of that layer. Figure \ref{fig:degree_reduction_algorithm} shows an overview of the heuristic algorithm and Algorithm \ref{alg:degree_reduction} describes the process of compressing an NDPNN by means of layer-wise polynomial degree reduction.
$F$ represents an NDPNN model pre-trained on a given training set $T$ such that $F(T)$ is the output of $F$ given $T$. Note that given an NPDNN model $F$, one can access the output of its layer $l$ by using $F_l$. The objective is to create another NDPNN model by reducing the polynomials of each layer of the pre-trained NDPNN model $F$ to the lowest possible polynomial degree such that the reduced model is faster, has less parameters, and performs as well as the pre-trained NDPNN model $F$ within a certain tolerance on the training set. In fact, this process is similar to decision tree pruning \cite{theoretical:pruning} in that it aims to reduce the complexity of the model without reducing its predictive capacity. To achieve that, we use the fact that, a pre-trained model will always output values in a certain interval on the training set, and that any polynomial of a given degree can be well-approximated by a polynomial of a lower degree on a symmetric interval using the polynomial degree reduction formula in \cite{intro:reduction}. The algorithm needs a performance evaluation function $\epsilon$ which takes a model, a dataset and its labels $T_{true}$ as input and needs a reduction tolerance $\epsilon_{0}$ which stops the algorithm when the performance evaluation of the reduced model is below $\epsilon_{0}$. $\epsilon$ can be any performance evaluation metric which outputs a higher score when the performance of the model is better, such as the accuracy or the negative mean squared error. The algorithm starts by initializing the reduction of each layer $R$ to $0$ and by determining the smallest symmetric interval $[-A_l,A_l]$ where the values of the input of each layer $l$ are bounded. Then, it goes through each layer $l'$, it creates a copy $\tilde{F}$ of the initial model $F$, and reduces every layer's degree $D_l$ of the copy $\tilde{F}$ by the last degree reduced $R$ on the symmetric interval $[-A_l,A_l]$, except for layer $l'$ which has its degree reduced by $R_{l'}+1$ on the symmetric interval $[-A_{l'},A_{l'}]$ as expressed by the instruction $\tilde{D}_l\gets D_l-\left(R_{l}+\mathds{1}_{\{l'\}}(l)\right)$ where $\mathds{1}$ is the indicator function. The weights of each layer of the initial model copy $\tilde{F}$ are therefore replaced during this process by reduced weights calculated via the polynomial degree reduction formula \cite{intro:reduction}. If a layer $l'$ has been reduced to the degree $1$, the algorithm does not attempt to reduce it further and ignores it by assigning a nil performance $P_{l'}=0$. It then evaluates the copy $\tilde{F}$ using the performance evaluation function $\epsilon$ and stores the score $P_{l'}$ in a list. After performing these steps for every layer $l'$ in the model, the algorithm determines the layer $\tilde{l}$ whose reduction impacted the performance of the model the least and increases its reduction $R_{\tilde{l}}$ by 1. These steps are repeated until all the layers are reduced to a degree of $1$ or until the best performance of the current reduction is below $\epsilon_{0}$. Following that, the algorithm creates a final copy $\tilde{F}$ of the model $F$ and reduces the degree of each layer $l$ according to the reduction limit $R_l$ determined in the previous steps. The algorithm then returns the most reduced model within the limit of $\epsilon_{0}$. Although the algorithm is computationally demanding, it allows the minimization of the polynomial degrees of every layer of a pre-trained NDPNN model with little to no loss in its prediction performance. This has the advantage of creating NDPNN models that solve highly complex tasks for a relatively low computational and spatial cost, which may prove necessary for the deployment in memory/compute-bound environments such as embedded systems.
\begin{figure}[!htpb]
\centering
\resizebox{0.6\linewidth}{!} 
{
\begin{algorithm2e}[H]
\caption{NDPNN layer-wise polynomial degree reduction}\label{alg:degree_reduction}
\KwData
{\newline
	$\bullet$ $L, N_l, D_l,W_{id}^{(l)}, b_i^{(l)}, \forall l\in[\![1,L]\!],\forall (i,d)\in[\![1,N_l]\!]\times[\![1,D_l]\!]$\newline 
	$\bullet$ Trained model $F$\newline
	$\bullet$ Training set $T$ and training labels $T_{true}$\newline
	$\bullet$ Evaluation function $\epsilon(F, T, T_{true})$\newline
	$\bullet$ Reduction tolerance $\epsilon_{0}$\newline
	
}
\KwResult
{
	\newline
	$\bullet$ $\tilde{D}_l,\tilde{W}_{id}^{(l)}, \tilde{b}_i^{(l)}, \forall l\in[\![1,L]\!],\forall (i,d)\in[\![1,N_l]\!]\times[\![1,\tilde{D}_l]\!]$\newline
	$\bullet$ Reduced model $\tilde{F}$
}
\SetAlgoLined\DontPrintSemicolon

	$A\gets(0,...,0)_L$ \;
	$R\gets(0,...,0)_L$ \;
	$P\gets(\epsilon_{0},...,\epsilon_{0})_L$ \;
	$A_1\gets\max\lvert T\rvert$\;
	\For{$l\in[\![2,L]\!]$}
	{
		$A_l\gets\max\left(\left\lvert F_{l-1}(T)\right\rvert\right)$\;		
	}
	
	$\tilde{l}\gets 0$\;
	\While{$P_{\tilde{l}}\geq\epsilon_{0} \land \underset{l\in[\![1,L]\!]}{\max} (D_l-R_l)>1$}
	{
	\For{$l'\in[\![1,L]\!]$}
	{
	$\tilde{F}\gets F$\;
   	\For{$l\in[\![1,L]\!]$}
	{	
		$\tilde{D}_l\gets D_l-\left(R_{l}+\mathds{1}_{\{l'\}}(l)\right)$\;	
		\eIf{$\tilde{D}_l\geq1$}
		{
			
			\For{$i\in[\![1,N_l]\!]$}
			{
				$\left(\left(\tilde{W}_{id}^{(l)}\right)_{[\![1,\tilde{D}_l]\!]}, \tilde{b}_i^{(l)}\right)\gets$reduce\_degree\_to\_n$\left(\left(W_{id}^{(l)}\right)_{[\![1,D_l]\!]}, b_i^{(l)}, D_l, \tilde{D}_l,A_l\right)$\;
				$\tilde{F}\gets$replace\_neuron\_weights$\left(\tilde{F}, \tilde{b}_i^{(l)}, \left(\tilde{W}_{id}^{(l)}\right)_{[\![1,\tilde{D}_l]\!]},\tilde{D}_l,i,l\right)$\;
			}			
		}
		{
			$P_{l'}\gets0$\;
			Break the innermost For loop\;
			
		}
			
	}
	$P_{l'}\gets \epsilon(\tilde{F},T, T_{true}) * \left(1-\mathds{1}_{\{0\}}(P_{l'})\right)$	\;
	}
	$\tilde{l}\gets\underset{l\in[\![1,L]\!]}{argmax}P$\;
	\If{$P_{\tilde{l}}\geq\epsilon_{0}$}
	{
		$R_{\tilde{l}}\gets R_{\tilde{l}} + 1$\;		
	}
	}
	$\tilde{F}\gets F$\;
	\For{$l\in[\![1,L]\!]$}
	{	
		$\tilde{D}_l\gets D_l-R_{l}$\;
		\For{$i\in[\![1,N_l]\!]$}
		{
			$\left(\left(\tilde{W}_{id}^{(l)}\right)_{[\![1,\tilde{D}_l]\!]}, \tilde{b}_i^{(l)}\right)\gets$reduce\_degree\_to\_n$\left(\left(W_{id}^{(l)}\right)_{[\![1,D_l]\!]}, b_i^{(l)}, D_l, \tilde{D}_l,A_l\right)$\;
			$\tilde{F}\gets$replace\_neuron\_weights$\left(\tilde{F}, \tilde{b}_i^{(l)}, \left(\tilde{W}_{id}^{(l)}\right)_{[\![1,\tilde{D}_l]\!]},\tilde{D}_l,i,l\right)$\;
		}	
	}
\end{algorithm2e}
}
\end{figure}
\subsection{Variably Overlapping Time--Coherent Sliding Window}\label{chapter=plants:sub=VOTCSW}
\begin{figure}[!htpb]
\centering
\subcaptionbox{2D convolution on an image.}{\includegraphics[scale=0.15]{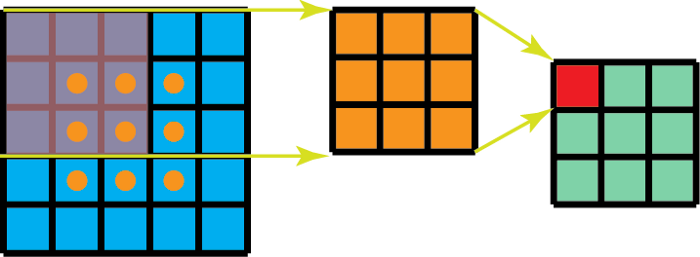}}\hspace{20pt}\subcaptionbox{3D convolution on a video.}{\includegraphics[scale=0.15]{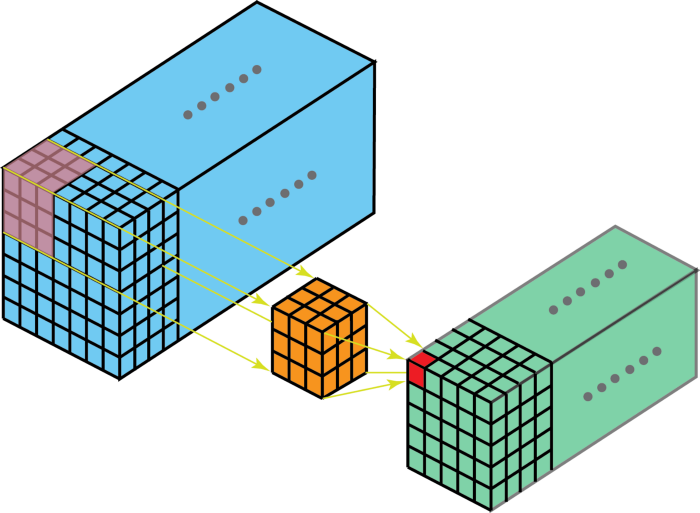}}
\vspace{10pt}
\caption[Difference between 2D convolution and 3D convolution.]{Difference between 2D convolution and 3D convolution \cite{2D_3D_conv}.}
\label{fig:2D_3D_conv}
\end{figure}

\begin{figure}[!htpb]
\centering
\includegraphics[scale=0.25]{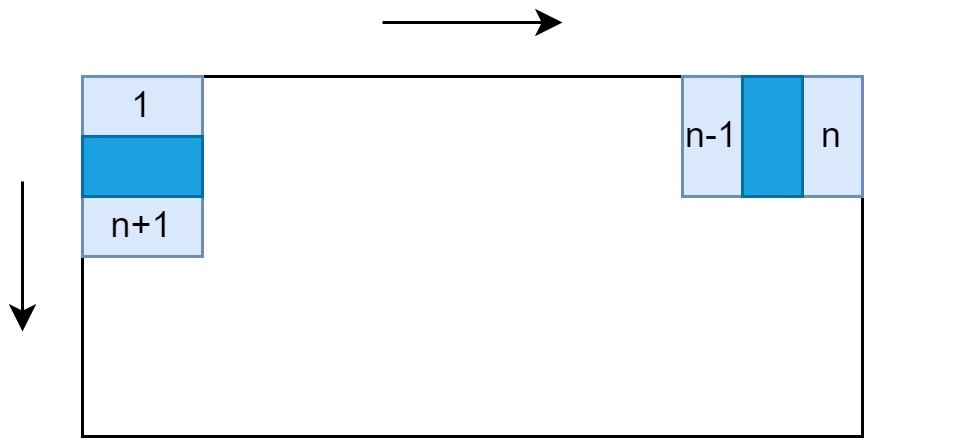}
\caption{Sliding windows on a rectangle representing an image. The arrows represent the sliding pattern and the dark blue color shows the overlap between two windows.}
\label{fig:sliding_window}
\end{figure}

The WSISCMC plant species classification dataset used in this work contains images with varying sizes which are not suitable for neural networks that only accept an input with a predetermined size. As a result, there is a need to transform these images into a representation with a fixed size. In most cases, shrinking the images to the smallest size present in the dataset is enough to train a network to produce very accurate results. However, image resizing comes at the cost of either losing important details that may be detrimental for the performance of the network when shrinking, or adding synthetic pixels when padding or magnifying. Thus, we created the Variably Overlapping Time--Coherent Sliding Window (VOTCSW) technique, which transforms each image, regardless of its size, to a 3-dimensional representation with fixed size $(h,w,M)$. The VOTCSW allows this representation to be interpreted as a video of size $(h,w)$ containing $M$ frames by ensuring that two consecutive frames are spatially correlated hence the ``Time--Coherent" term. Therefore, the 3-dimensional representation can be fed into 3DCNNs and 3DPNNs as a tensor of shape $(h,w,M,C)$ where $C$ is the number of channels present in the original image. Figure \ref{fig:2D_3D_conv} (adapted from \cite{2D_3D_conv}) shows the difference between a 2D convolution applied on an image and a 3D convolution applied on a video similar to the ones that can be created using the VOTCSW method. The difference resides in the way a filter is defined and the way it slides. To apply 2D convolution on an image, a 2D filter -- of size $3\times3$ in this example -- slides over the 2 dimensions that define an image, namely, height and width, to produce a feature map. To apply 3D convolution on a video, a 3D filter -- of size $3\times3\times3$ in this example -- slides over the 3 dimensions that define a video, namely, height, width and time, to produce a feature map.
The VOTCSW method is a safe alternative to resizing as it does not add nor remove any pixel from the original images, or at least, it minimizes the need to do so under certain conditions that will be discussed below. It is based on the sliding window technique (hence the ``Sliding Window" term) which is a powerful signal processing tool that is used to decompose a signal containing a high number of samples into small chunks called windows containing a smaller number of samples that can be processed faster. Consecutive windows overlap with a certain ratio $\alpha\in[0,1[$ in order to ensure a correlation between them. The classical use of this technique consists in determining a window size that ensures enough representative samples to be present in a single window and an overlap that allows better processing performance. However, there is no consideration as to how many windows are extracted for each different signal length. In constrast, our proposed VOTCSW method aims to extract exactly $M$ windows of a fixed size from any signal regardless of its length. This is achieved by calculating an overlap for each signal length, hence the Variably Overlapping term.
\newline\indent
Given an image of size $(H,W)$ and a desired 3-dimensional representation $(h,w,M)$, we define the following relationships:
\begin{flalign}\label{eq:H&W}
\begin{cases}
H &= h + N_h(1-\alpha)h\\
W &= w + N_w(1-\alpha)w\\
\end{cases},&&
\end{flalign}
where $\alpha\in[0,1[$ is the window overlap, $N_h$ is the number of windows that overlap with their predecessors and that are needed to cover the height of the image, and $N_w$ is the number of windows that overlap with their predecessors and that are needed to cover the width of the image. Following that, the total number of windows $M$ needed to cover the image in its entirety is
\begin{flalign}\label{eq:M}
M=(1+N_h)(1+N_w)=\left(\dfrac{H-h}{(1-\alpha)h}+1\right)\left(\dfrac{W-w}{(1-\alpha)w}+1\right).&&
\end{flalign}
Eq. (\ref{eq:M}) establishes a relationship between the overlap $\alpha$, the size of the image $(H,W)$ which is fixed, the size of the sliding window $(h,w)$ which is fixed and the total number of windows $M$ which is also fixed.  
\begin{proposition}\label{prop:alpha}
The value of $\alpha$ with respect to $H$, $h$, $W$, $w$ and $M$ is
\begin{flalign}\label{eq:alpha}
\alpha=1-\dfrac{((H-h)w+(W-w)h)+\sqrt{\Delta}}{2hw(M-1)},&&
\end{flalign}
where $\Delta=((H-h)w+(W-w)h)^2+4hw(H-h)(W-w)(M-1)$.
\end{proposition}
Refer to Appendix A for proof. Eq. (\ref{eq:alpha}) was determined from the fact that $\alpha<1$. However, $\alpha$ needs to be positive as well. Therefore, there is a need to determine a condition on the choice of $h$, $w$ and $M$ with respect to $H$ and $W$ to ensure the positivity of $\alpha$.
\begin{proposition}\label{prop:alpha_positive}
\begin{flalign*}
\alpha\geq0\iff hwM\geq HW.&&
\end{flalign*}
\end{proposition}
Refer to Appendix A for proof. Proposition \ref{prop:alpha_positive} implies that $h$, $w$ and $M$ can not be chosen arbitrarily. Furthermore, since $h$, $w$ and $M$ need to be fixed before transforming the images of the dataset, they have to verify this condition for every image of size $(H,W)$. As a result, we need to determine tighter conditions to be able to determine $h$, $w$ and $M$ consistently. Let $\beta$ be the aspect ratio of an image of size $(H,W)$ such that $W=\beta H$ and let $\gamma$ be the aspect ratio of a window of size $(h,w)$ such that $w=\gamma h$. We then derive from Eq. (\ref{eq:H&W})
\begin{flalign*}
W = w + (1-\alpha)wN_w\iff\beta H = \gamma h +(1-\alpha)\beta hN_w\iff N_w = \dfrac{\beta H-\gamma h}{(1-\alpha)\gamma h}.&&
\end{flalign*}
And since from Eq. (\ref{eq:H&W}), we have $N_h=\dfrac{H-h}{(1-\alpha)h}$, we deduce by dividing $N_w$ by $N_h$ that $N_w=\dfrac{\beta H-\gamma h}{\gamma(H-h)}N_h$ and thus
\begin{flalign}\label{eq:M_N_h}
M=(N_h+1)\left(\dfrac{\beta H-\gamma h}{\gamma(H-h)}N_h+1\right).&&
\end{flalign}
Since $M$ is an integer constant, and $N_h$ is an integer constant, $\dfrac{\beta H-\gamma h}{\gamma(H-h)}N_h$ should also be an integer constant. As a result, we can define a positive constant $p$ such that $p= \dfrac{\beta H-\gamma h}{\gamma(H-h)}$. The only variables in $p$ are $\beta$ and $H$ since they depend on the image being processed. $\gamma$ and $h$ are not supposed to change with the size of the image being processed. Consequently, we can write
\begin{flalign}\label{eq:p}
p = \dfrac{\beta H-\gamma h}{\gamma(H-h)}\iff \gamma(H-h)p+\gamma h=\beta H\iff H(\beta-\gamma p)=\gamma h(1-p).&&
\end{flalign}
This means that when $p\neq1$ and $\beta\neq\gamma p$, we have $H=\dfrac{\gamma h (1-p)}{\beta-\gamma p}$. This implies that, in order for $M$ to be a valid integer, every image height has to be resized to $H$, which contradicts the purpose of the VOTCSW. Therefore, we consider the case where $\beta=\gamma p$. We can deduce from Eq. (\ref{eq:p}) that, under this condition, $p$ will be equal to $1$ and $\beta=\gamma$. As a result, the aspect ratio of each image should be equal to that of the sliding window for $M$ to be a valid integer, which is a simpler condition than the previous one. In the following, we will only consider the case where $\beta=\gamma$ since the dataset used in this work contains images that have the same aspect ratio. When $\beta=\gamma$, $M=(N_h+1)^2$ which means that $M$ should be a square number. This is yet another condition on how to choose $M$ and this narrows down the possibilities even further. Under this assumption, the calculation of $\alpha$ can be simplified.
\begin{proposition}\label{prop:alpha_simple}
When $\beta=\gamma$, $\alpha$ only depends on $M$, $H$ and $h$ and its value is
\begin{flalign}\label{eq:alpha_simple}
\alpha=\dfrac{\sqrt{M}h-H}{h(\sqrt{M}-1)}.&&
\end{flalign}
\end{proposition}
Refer to Appendix A for proof. When extracting the windows from an image, we should limit the overlap $\alpha$ to not reach its extremum in order to obtain consistent windows. Therefore, we impose on $\alpha$ two limits $\alpha_{min}$ and $\alpha_{max}$ such that $0\leq\alpha_{min}\leq\alpha\leq\alpha_{max}<1$. Given $H_{max}$, the height of the biggest image in the dataset and $H_{min}$, the height of the smallest image in the dataset, we can formulate a condition on $h$.
\begin{proposition}\label{prop:h}
The height $h$ of the sliding window has to verify the following condition:
\begin{flalign}\label{eq:h}
\dfrac{H_{max}}{\sqrt{M}-\alpha_{min}(\sqrt{M}-1)}\leq h\leq \dfrac{H_{min}}{\sqrt{M}-\alpha_{max}(\sqrt{M}-1)}.&&
\end{flalign}
\end{proposition}
Refer to Appendix A for proof. Although a condition on the choice $h$ is important, the parameter that will most likely be chosen first when using the VOTCSW technique is $M$. However, $\alpha_{min}$ and $\alpha_{max}$ are as important as $M$ because they specify the maximum and minimum amount of correlation between two consecutive windows. Hence, a condition on their choice is also important. 
\begin{theorem}\label{theorem:condition on the parameters}
The parameters $M$, $\alpha_{min}$ and $\alpha_{max}$ can be determined in 6 different ways. For each way, there are conditions that need to be verified in order to extract the windows correctly.
\newline
$\bullet$ When determining $M$ then $\alpha_{min}$ then $\alpha_{max}$, the following conditions apply:
\begin{flalign*}
\begin{cases}
\sqrt{M}&\geq\dfrac{H_{max}}{H_{min}}\\
\alpha_{min}&\leq \dfrac{H_{max}}{H_{min}}+\left(1-\dfrac{H_{max}}{H_{min}}\right)\dfrac{\sqrt{M}}{\sqrt{M}-1}\\
\alpha_{max}&\geq\left(1-\dfrac{H_{min}}{H_{max}}\right)\dfrac{\sqrt{M}}{\sqrt{M}-1}+\dfrac{H_{min}}{H_{max}}\alpha_{min}\\
\end{cases}&&
\end{flalign*}
$\bullet$ When determining $M$ then $\alpha_{max}$ then $\alpha_{min}$, the following conditions apply:
\begin{flalign*}
\begin{cases}
\sqrt{M}&\geq\dfrac{H_{max}}{H_{min}}\\
\alpha_{max}&\geq\left(1-\dfrac{H_{min}}{H_{max}}\right)\dfrac{\sqrt{M}}{\sqrt{M}-1}\\
\alpha_{min}&\leq \dfrac{H_{max}}{H_{min}}\alpha_{max}+\left(1-\dfrac{H_{max}}{H_{min}}\right)\dfrac{\sqrt{M}}{\sqrt{M}-1}\\
\end{cases}&&
\end{flalign*}
$\bullet$ When determining $\alpha_{min}$ then $M$ then $\alpha_{max}$, the following conditions apply:
\begin{flalign}\label{eq:alpha_min_m}
\begin{cases}
\sqrt{M}&\geq\dfrac{H_{max}-H_{min}\alpha_{min}}{(1-\alpha_{min})H_{min}}\\
\alpha_{max}&\geq\left(1-\dfrac{H_{min}}{H_{max}}\right)\dfrac{\sqrt{M}}{\sqrt{M}-1}+\dfrac{H_{min}}{H_{max}}\alpha_{min}\\
\end{cases}&&
\end{flalign}
$\bullet$ When determining $\alpha_{min}$, then $\alpha_{max}$ then $M$, the following conditions apply:
\begin{flalign}\label{eq:alpha_min_alpha_max}
\begin{cases}
\alpha_{max}&\geq1-(1-\alpha_{min})\dfrac{H_{min}}{H_{max}}\\
\sqrt{M}&\geq\dfrac{H_{max}\alpha_{max}-H_{min}\alpha_{min}}{H_{min}(1-\alpha_{min})-H_{max}(1-\alpha_{max})}\\
\end{cases}&&
\end{flalign}
$\bullet$ When determining $\alpha_{max}$ then $M$ then $\alpha_{min}$, the following conditions apply:
\begin{flalign*}
\begin{cases}
\alpha_{max}&\geq1-\dfrac{H_{min}}{H_{max}}\\
\sqrt{M}&\geq\dfrac{H_{max}\alpha_{max}}{H_{min}-(1-\alpha_{max})H_{max}}\\
\alpha_{min}&\leq \dfrac{H_{max}}{H_{min}}\alpha_{max}+\left(1-\dfrac{H_{max}}{H_{min}}\right)\dfrac{\sqrt{M}}{\sqrt{M}-1}\\
\end{cases}&&
\end{flalign*}
$\bullet$ When determining $\alpha_{max}$ then $\alpha_{min}$ then $M$, the following conditions apply:
\begin{flalign*}
\begin{cases}
\alpha_{max}&\geq1-\dfrac{H_{min}}{H_{max}}\\
\alpha_{min}&\leq 1-\dfrac{H_{max}}{H_{min}}(1-\alpha_{max})\\
\sqrt{M}&\geq\dfrac{H_{max}\alpha_{max}-H_{min}\alpha_{min}}{H_{min}(1-\alpha_{min})-H_{max}(1-\alpha_{max})}\\
\end{cases}&&
\end{flalign*}
\end{theorem}
Refer to Appendix A for proof. Theorem \ref{theorem:condition on the parameters} shows that in 2 particular cases described by Eqs. (\ref{eq:alpha_min_m}) and (\ref{eq:alpha_min_alpha_max}), when $\alpha_{min}$ is determined first, there is no constraint on its value other than that it should be positive and lower than 1. These 2 cases should be preferred over the other more constrained ones when choosing the values of $\alpha_{min}$, $\alpha_{max}$ and $M$. The theorem also ensures the choice of well defined parameters, given $H_{min}$ and $H_{max}$ only. Nevertheless, in some datasets, the difference between $H_{min}$ and $H_{max}$ is considerable and may lead to the choice of a high number of windows $M$, a maximum overlap $\alpha_{max}$ close to 1, or a minimum overlap $\alpha_{min}$ close to 0. Therefore, one can also arbitrarily choose the parameters that are deemed appropriate and then choose a maximum height $\tilde{H}_{max}$ and a minimum height $\tilde{H}_{min}$ such that:
\begin{flalign}\label{eq:H_max/H_min}
\dfrac{\tilde{H}_{max}}{\tilde{H}_{min}}\leq \dfrac{\sqrt{M}-\alpha_{min}(\sqrt{M}-1)}{\sqrt{M}-\alpha_{max}(\sqrt{M}-1)}.&&
\end{flalign}
After performing this choice, each image in the dataset whose height exceeds $\tilde{H}_{max}$ should be shrinked to $\tilde{H}_{max}$, and each image whose height is less than $\tilde{H}_{min}$ should be padded or magnified to $\tilde{H}_{min}$. Although this defeats the purpose of the VOTCSW method, it is still better than resizing all the images in the dataset since the images whose heights are within $[\tilde{H}_{min},\tilde{H}_{max}]$ will remain the same. Using Proposition \ref{prop:alpha_simple}, Proposition \ref{prop:h} and Theorem \ref{theorem:condition on the parameters} enables any image whose height is between the limits $H_{min}$ and $H_{max}$ and whose aspect ratio $\beta$ is the same as that of the sliding window $\gamma$ to be transformed in a fixed 3-dimensional representation $(h,\gamma h,M)$. However, the order of the windows in the 3-dimensional sequence should not be arbitrary as it should ensure a correlation between every two consecutive windows. The VOTCSW method is based on the sliding window technique which is widely used on 1-dimensional signals because the inter-window correlation is always ensured. However, this is no longer guaranteed for higher dimensional signals such as images as shown in Figure \ref{fig:sliding_window} which describes the sliding window technique on an image represented by a rectangle where the the light blue color represents a window a the dark blue color shows the overlap between two windows. As represented by the arrows, the window slides from left to right and resumes to the initial position when it reaches the boundaries of the image. It then slides one step downward and continues sliding from left to right until it covers the whole image. This sliding pattern does not ensure that every two consecutive windows are correlated and this can be noticed in the aforementioned figure where window $n$ and window $n+1$ are totally separated.
Nevertheless, this usually does not matter in machine learning applications where each window can be treated independently and the result is determined by aggregating the results obtained on each individual window. However, the VOTCSW method consists in creating a causal 3-dimensional representation that is analogous to a video and which can be processed as a video in the sense that there is a time coherence between two consecutive windows meaning that one necessarily appears before the other and is spatially correlated with it. To ensure that, we define, given a matrix $I$ of size $H\times W$ representing an image, the following sequences describing a time--coherent left-to-right, top-to-bottom sliding pattern:
\begin{flalign}\label{eq:sliding_pattern}
\begin{split}
&\forall n\in[\![0,M-1]\!],\\
&\delta_n = \left\lfloor\dfrac{n}{\sqrt{M}}\right\rfloor,\\
&a_n=(1-\alpha)h\delta_n,\\
&b_n=a_n+h,\\
&c_n=\left(\left(1-(-1)^{\delta_n}\right)\dfrac{\sqrt{M}-1}{2}+(-1)^{\delta_n}\left(n-\delta_n\sqrt{M}\right)\right)(1-\alpha)\gamma h,\\
&d_n = c_n+\gamma h,\\
&\mathcal{W}_n = \begin{bmatrix} 
    I_{a_nc_n} & I_{a_nc_{n+1}} & \dots \\
    \vdots & \ddots & \\
    I_{b_nc_n} &        & I_{b_nd_n} 
    \end{bmatrix},
\end{split}&&
\end{flalign}
where $\mathcal{W}_n$ is the n-th window extracted from the image $I$, $h$ is a constant determined using Theorem \ref{theorem:condition on the parameters} and Proposition \ref{prop:h}, and $\alpha$ is calculated using Proposition \ref{prop:alpha_simple}. This pattern is a modification of the one described in Figure \ref{fig:sliding_window} such that when the window slides to the right edge of the rectangle, it slides downward by a step, and slides back to the left until reaching the left edge before it slides downward again and slides back to the right edge. As a result, this pattern ensures the time--coherence of the 3-dimensional representation created by the VOTCSW method. Another property of the VOTCSW method is that it performs an oversampling of the pixels due to the window overlap that enables the window to cover the same pixel more than once.
\begin{proposition}\label{prop:oversampling}
The maximum oversampling factor for a pixel in an image that is processed with the VOTCSW method with an overlap $\alpha$ is $\dfrac{1}{(1-\alpha)^2}$.
\end{proposition}
Refer to Appendix A for proof. Proposition \ref{prop:oversampling} implies that $\alpha_{min}$ and $\alpha_{max}$ are means to control the maximum oversampling factor of a pixel. Moreover, it implies that, the smaller the image, the more its pixels will be oversampled which will definitely alter the class distribution for a classification problem if the size distribution in each class is uneven. This may prove useful in certain cases of image classification where the images representing the least represented class happen to be the smallest in size. Finally, the VOTCSW method can be summarized in the following steps:
\begin{enumerate} 
\item[1.] Ensure that the image dataset has a single aspect ratio $\beta$ and determine $H_{max}$ and $H_{min}$.
\item[2.] Choose $\alpha_{min}$, $\alpha_{max}$, $M$ and $h$ as recommended in Theorem \ref{theorem:condition on the parameters} and Proposition \ref{prop:h}. Define $\gamma=\beta$.
\item[3.] Choose a time--coherent sliding pattern such as the one described in Eq. (\ref{eq:sliding_pattern}) and use it for each image in the dataset.
\end{enumerate}
\section{Experiments and results}\label{chapter=plants:section=experiments}
In order to produce meaningful results and to reliably choose a model over the other, the framework shown in Figure \ref{fig:plants_framework} was designed. This section details and discusses each block in the diagram and provides an in-depth analysis of the results obtained on the WSISCMC dataset.
\subsection{Preprocessing}
\begin{figure}[!htpb]
\centering
\subcaptionbox{Training set and test set distribution.}{\includegraphics[scale=0.049]{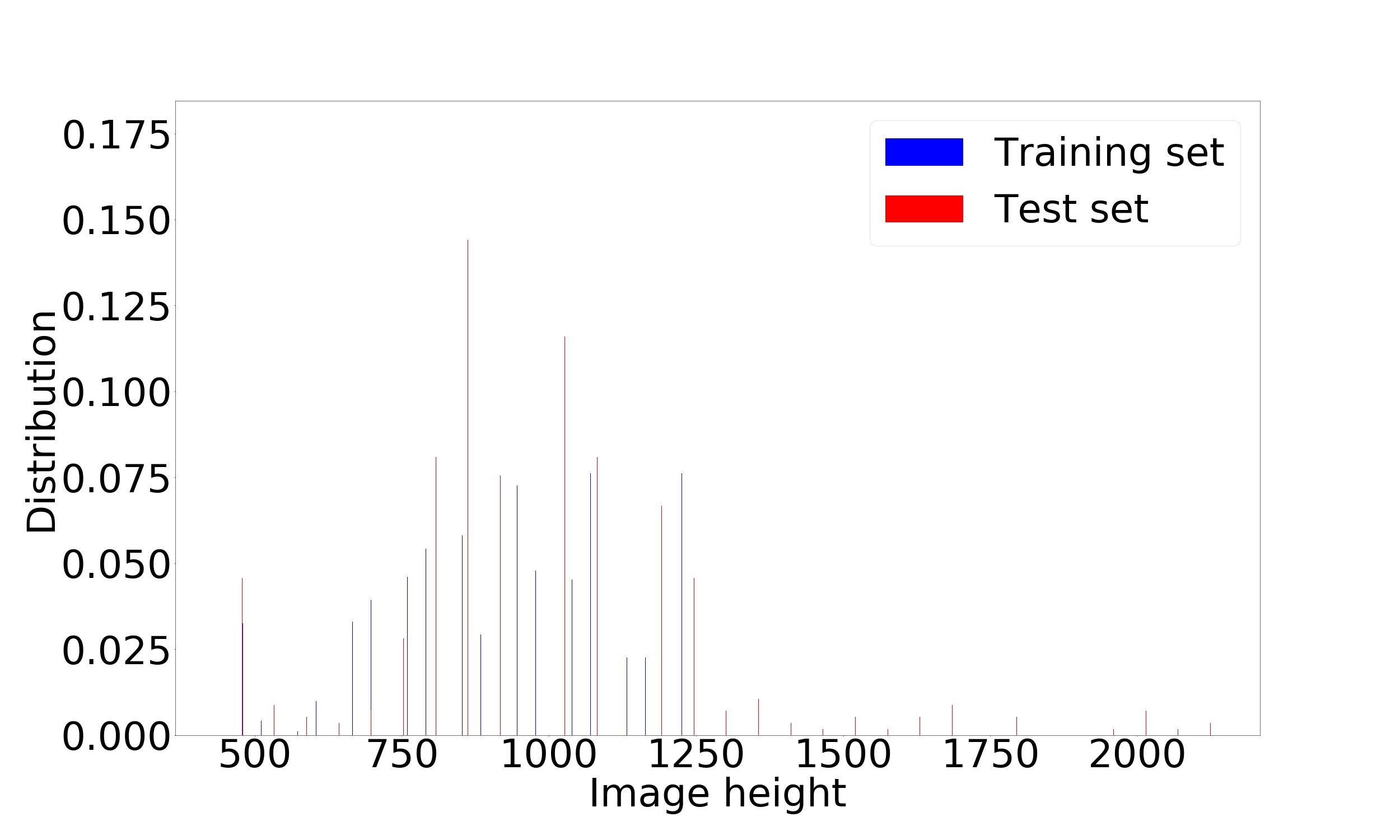}}\subcaptionbox{Reworked dataset distribution.}{\includegraphics[scale=0.049]{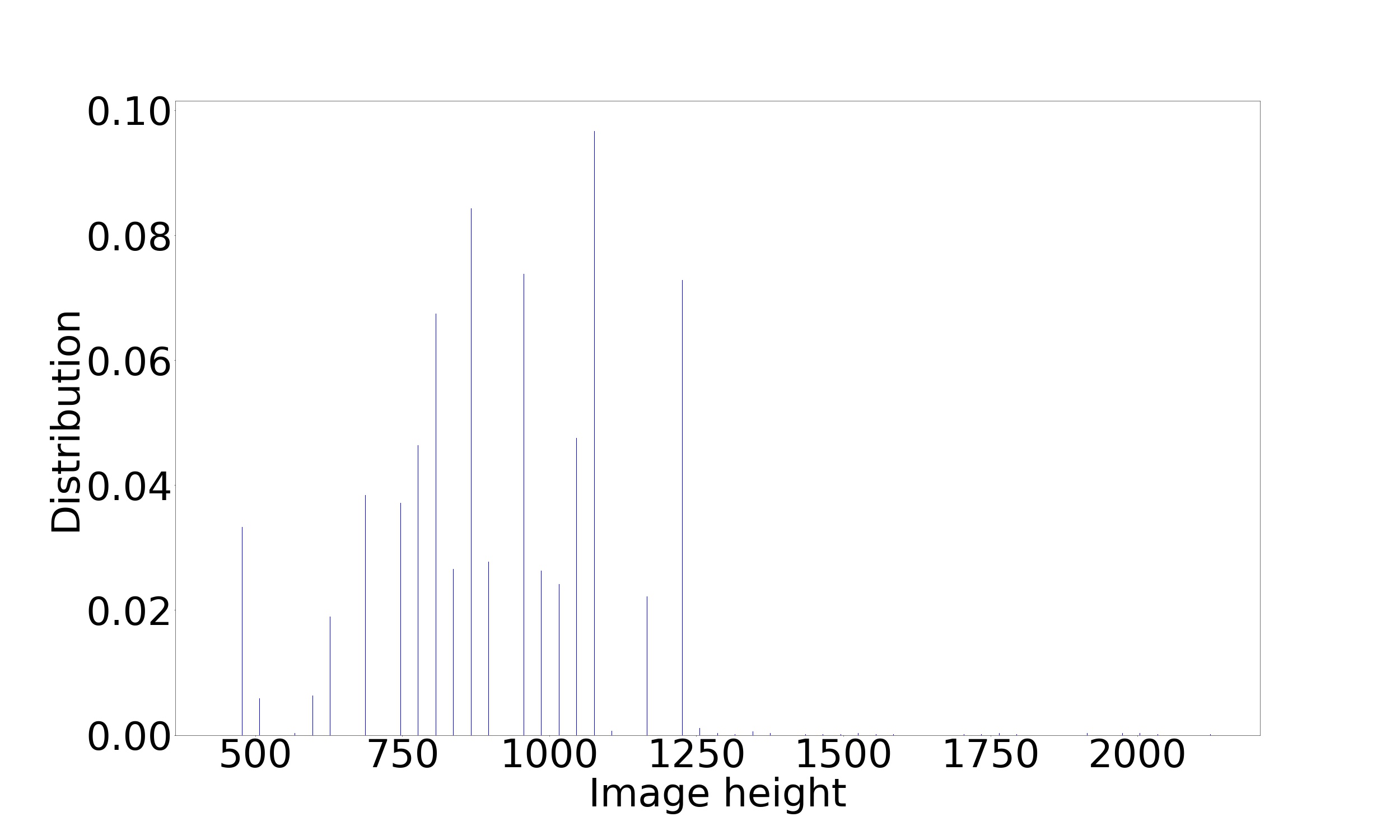}}
\vspace{10pt}
\caption{Size distributions for the ``Canola" class in the training set, the test set and the reworked dataset.}
\label{fig:size_distribution}
\end{figure}

The WSISCMC plant species classification dataset that is used in this work was mainly constructed to specifically overcome the limitations of the current state-of-the-art datasets used in machine learning which consist in a lack of sufficient variety and quantity. It contains $38,680$ high-quality square photos with different sizes of 8 different species of plants taken from different angles, and at various growth stages, which can theoretically enable a well-trained classification model to recognize plants at any stage in their growth. Moreover, the plants are potted and placed in front of a uniform background that can easily be substituted with field images, for example. There are also images which are taken with a mobile phone in various backgrounds to enable trained models to be tested on unprocessed images. In \cite{intro:EAGL-I}, it was shown that the WSISCMC dataset allows the production of a reliable binary classifier that differentiates between grasses and non-grasses. However, the dataset was not used to train a more complex model of plant species classification. A first attempt to evaluate the difficulty of this task on this dataset was to create a baseline 2DCNN that takes images resized to $224\times224$ and attempts to predict the species of the plant present in the dataset. The tested accuracy of that baseline model was $49.23\%$, which was mediocre. 
\newline\indent
It was then determined, after an analysis of the dataset, that there was a significant imbalance in the class distribution. Moreover, this distribution was radically different between the training set and the test set. Furthermore, the size distribution was also very different between the training set and test set, such that the maximum image size in the training set was $1226\times1226$ and the maximum image size in the test set was $2346\times2346$. In addition, the distribution of size per class also varied between the training set and the test set. Therefore, the dataset was entirely redistributed into a training set and a test set that have the same class distribution and the same size per class distribution while keeping the same train-test ratio as the initial dataset. Table \ref{table:plants_distribution} shows the initial training set and test set class distributions as well as the class distribution of the reworked (redistributed) dataset.
\begin{table}[!htpb]
\centering
\resizebox{\columnwidth}{!}{
\begin{tabular}{llll}
Species&Training set distribution&Test set distribution&Reworked distribution\\\hline
Smartweed&0.03&0.14&0.04\\\hline
Yellow Foxtail&0.10&0.22&0.11\\\hline
Barnyard Grass&0.25&0.12&0.23\\\hline
Wild Buckwheat&0.12&0.14&0.12\\\hline
Canola&0.19&0.14&0.19\\\hline
Canada Thistle&0.14&0.14&0.14\\\hline
Dandelion&0.14&0.10&0.13\\\hline
Wild Oat& 0.03& 0&0.03\\\hline
\textbf{Total}&1&1&1\\\hline
\end{tabular}
}
\caption{Training set, test set and reworked distributions of the 8 species in the dataset.}
\label{table:plants_distribution}
\end{table}
We notice that the ``Wild Oat" class is not even present in the test set and that the ``Smartweed" and ``Yellow Foxtail" classes are the most represented classes in the test set, and the least represented ones in the training set. Figure \ref{fig:size_distribution}.(a) shows the size distribution of the ``Canola" class in the training set and the test set whereas Figure \ref{fig:size_distribution}.(b) shows the size distribution of the ``Canola" class in the reworked dataset. We notice that the size distribution of the ``Canola" class in the training set is widely different from the test set and that the reworked dataset ensures that every size is present with the same proportion in the training set and the test set. 
\newline\indent
These factors, as well as the aforementioned ones, explain why the baseline model failed to accurately determine the species of the images it processed. Nevertheless, after reworking the dataset, the baseline model achieved an accuracy of $95.1\%$. Although this was a noticeable improvement, there was still room to achieve better accuracy with further preprocessing. One of which is to modify the bias initialization of the classifier's last layer to take into account the inherent data imbalance. Since the last layer uses a softmax activation function, we need to solve the following system of equations:
\begin{flalign*}
\forall k\in[\![1,N]\!], p_k = \dfrac{e^{b_k}}{\sum_{i=1}^{N}e^{b_i}},&&
\end{flalign*}
where $N$ is the number of classes, $p_k$ is the presence of class $k$ in the dataset, and $b_k$ is the bias in the neuron $k$ that will predict the probability of an image to belong to class $k$. This system is linear in $e^{b_k}$ and is easily solvable by noticing that $\dfrac{p_j}{p_k}=e^{b_j-b_k}$. This bias modification improved the accuracy of the baseline model by $1.58\%$ which reached $96.68\%$ accuracy. Consequently, we created 6 different versions of the same dataset in order to evaluate the gain of using the VOTCSW method with respect to resizing (shrinking, padding, or magnifying). The first three versions are produced by the VOTCSW method with the following parameters $\alpha_{min}=0.1$, $\alpha_{max}=0.9$ and $M=9$. These parameters impose that we shrink the images whose height is greater than $973$ to $973$. This value was determined using Eq. (\ref{eq:H_max/H_min}) for $H_{min}=418$ which represents the lowest size present in the dataset. The difference between the first three versions produced by the VOTCSW method is the sliding patterns which are horizontal (refer to Eq. (\ref{eq:sliding_pattern})), vertical and spiral, respectively. The images were all transformed to 3D tensors of size $348\times348\times9$ according to Eq. (\ref{eq:h}), meaning that they are equivalent to videos of size $348\times348$ containing $9$ frames. The fourth version and fifth version of the dataset consist in resizing the images to a size that is equivalent, in terms of the number of samples, to the size of the ``videos" generated by the VOTCSW method. This size is $\sqrt{348\times348\times9}=1044$ and the images that are larger than $1044$ are shrinked to $1044$ whereas the ones that are smaller than $1044$ are zero-padded to that size in the fourth version, and magnified to that size in the fifth version. The sixth version of the dataset consists in shrinking all the images to a size of $224\times224$ which is suitable for using well-known architectures such as ResNet or Inception. The 6 versions of the dataset are referred to as WSISCMC--H, WSISCMC--V, WSISCMC--S, WSISCMC--1044P, WSISCMC--1044M, and WSISCMC--224 following the order in which they were introduced above.
\subsection{Model development}
\begin{figure}[!htpb]
\centering
\includegraphics[scale=0.1]{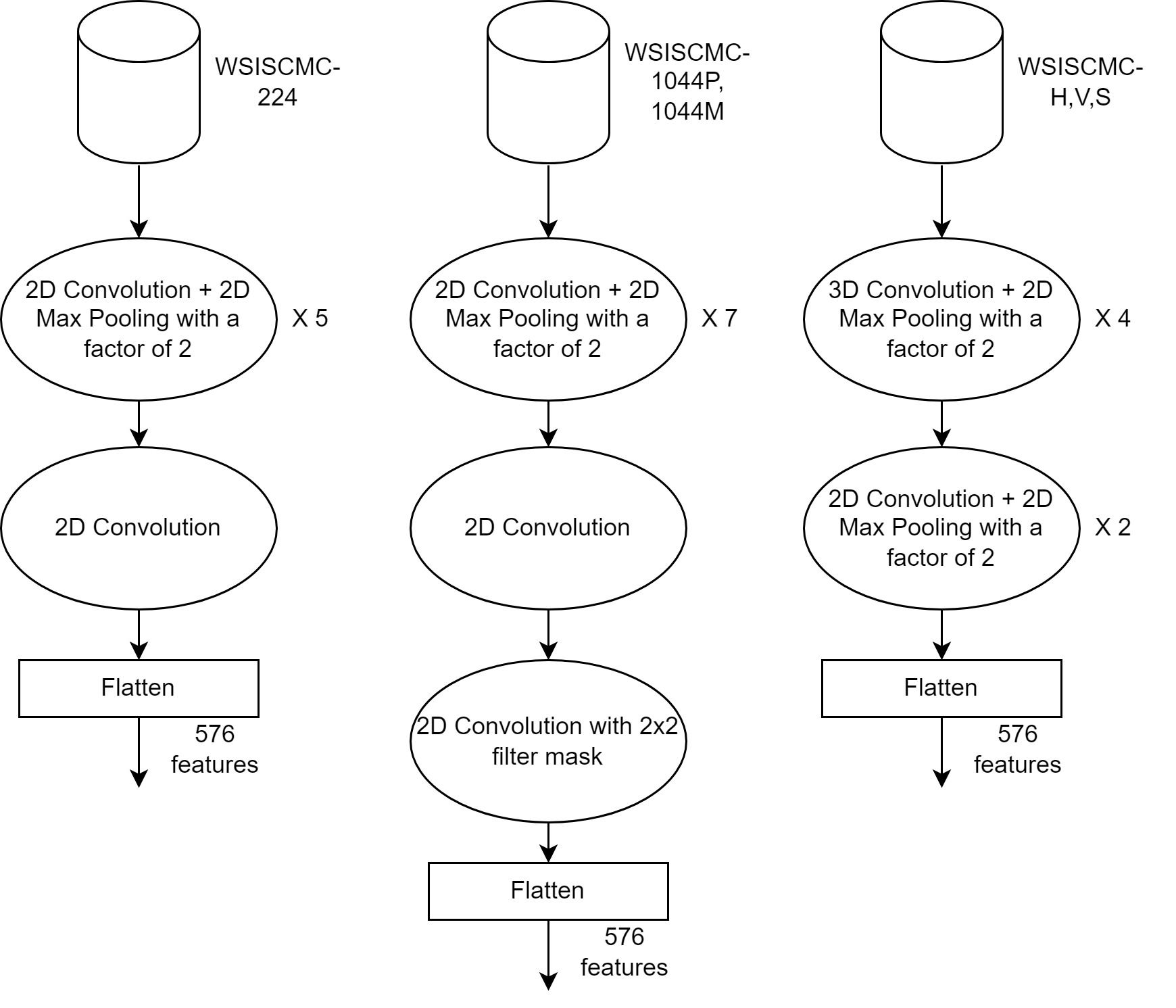}
\caption{Network architecture for each version of the WSISCMC dataset.}
\label{fig:network_architecture}
\end{figure}

After performing the preprocessing described above, we created multiple variations for each kind of model. As shown in Figure \ref{fig:plants_framework}, we created an incremental model development framework which allows us to study the influence of every technical addition to basic 2DCNNs, hence, to perform an ablation study \cite{intro:ablation}. In fact, we start by creating 3 different 2DCNNs on the shrinked, padded and magnified versions of the dataset respectively, in order to determine the difference between the 3 resizing approaches. Then, we create 3 3DCNNs with different sliding movements to determine the difference between using the VOTCSW and using conventional resizing with 2DCNNs. We also ensure that the 3DCNNs have the same number of parameters as the 2DCNNs to better differentiate between the representational power of the VOTCSW and that of conventional resizing. Indeed, since 2DCNNs and 3DCNNs are fundamentally the same on an operational level and only differ on the representational level as shown in Figure \ref{fig:2D_3D_conv}, comparing the performances of 2DCNNs and 3DCNNs that have the same number of parameters on the same task results in comparing their respective input representation efficiencies. Following that, we extend the best 2DCNN and the best 3DCNN to NDPNNs which allows us to evaluate the gain of using NDPNNs over NDCNNs. Finally, we apply the layer-wise degree reduction heuristic to determine the efficiency of polynomial degree reduction applied on NDPNNs. All the networks used share in common a $3\times3$ ($3\times3\times3$ for 3D) filter mask size for the convolution layers, an initial layer composed of $32$ neurons, a last feature extraction layer composed of $64$ neurons, the use of ReLU \cite{relu} as activation function, the shape of the output of their feature extractor which is $576$ features and their densely connected layers which are composed of 2 layers, one containing $128$ neurons, followed by one containing $8$ neurons corresponding to the $8$ classes of the dataset with a softmax activation. For each dataset version, there is a predetermined depth for the networks created as shown in Figure \ref{fig:network_architecture}. We performed a grid search on the number of neurons for the 4 penultimate feature extraction layers of every 2D network with 10-fold cross validation and we ensure that the validation set always has the same class distribution and size distribution as the train set. The possible values used for the number of neurons in these layers were $16$, $32$ and $64$ to keep the experiments feasible to be completed in a reasonable amount of time. The layers that were not searched were composed of $64$ neurons by default. This search allowed us to select the best architecture among $81$ variations for each of the 3 2D dataset versions, WSISCMC--224, WSISCMC--1044P and WSISCMC--1044M. 
\newline\indent
Following that, for each 2D version of the dataset, a network was created and trained with the best determined architecture 10 times with different initial weights and evaluated on its corresponding test set such that only the best one is chosen. Then, the 2DCNN architecture of the best network among the one trained on the WSISCMC--1044P and the one trained on the WSISCMC--1044M was chosen to create a 3DCNN architecture that has the same number of parameters as the chosen architecture. This is to ensure a fair evaluation of the effect of the VOTCSW method on the performance of the 3DCNN. Indeed, if a 3DCNN with more parameters than a 2DCNN trained on WSISCMC--1044P or WSISCMC--1044M achieves better results than the 2DCNN, then this might be due to the difference in the number of parameters, and not the difference in data representation. Since 2DCNNs and 3DCNNs perform the same fundamental operation, if both have the same number of parameters and one of them performs better than the other, then this is most likely due to the difference in data representation. Naturally, the weight initialization also plays a role in this difference in performance, therefore, we always train the same network 10 times with different initial weights and choose the one that achieves the best accuracy on the test set. The architecture of the 3DCNN that is calculated from the best 2DCNN architecture has to be consistent with the above description in the sense that the initial layer contains $32$ neurons, and the last feature extraction layer contains $64$ neurons. We assume that all the inner layers of the 3DCNN have the same number of neurons $N$. If we denote the number of parameters of the feature extractor of the best network trained on WSISCMC--1044P or WSISCMC--1044M by $N_{1044}$, the receptive field of the 3D convolution layers by $R_3$ ($27$ in this case), and the receptive field of the 2D convolution layers by $R_2$ ($9$ in this case), then the number of neurons $N$ of each inner layer of the 3DCNN is determined by the following equation:
\begin{flalign*}
3\times32\times R_3 + 32R_3 N + 2 R_3 N^2 + R_2N^2+ 64R_2N + 32 + 64 + 4N=N_{1044},&&
\end{flalign*}
which is equivalent to
\begin{flalign}\label{eq:N}
(2R_3+R_2)N^2+4(8R_3+16R_2+1)N+96(R_3+1)-N_{1044}=0.&&
\end{flalign}
This second degree equation with unknown $N$ has a unique positive solution because $96(R_3+1)-N_{1044}$ is negative and the other coefficients are positive. Since $N$ has to be an integer, the determined solution is rounded before it is used.
\newline\indent
The resulting architecture was then used to create and train 3DCNNs on WSISCMC--H, WSISCMC--V and WSISCMC--S. The best 3DCNN network and the best 2DCNN network were then selected to be extended to NDPNNs. Consequently, each convolution layer of each network was changed to an NDPNN layer with a degree $7$, such that a 2DPNN and a 3DPNN were trained and evaluated on their proper respective datasets. The degree $7$ was chosen because we observed that most usual functions could be well approximated by polynomials of degree $7$ on the interval $[-1,1]$. Finally, both of the 2DPNN and the 3DPNN were reduced using the layer-wise degree reduction heuristic described in Algorithm \ref{alg:degree_reduction} with $0$ tolerance to gain computational efficiency and reduce memory usage without sacrificing their accuracy. We also fine-tuned a ResNet50V2, an InceptionV3 and a Xception network on the WSISCMC--224 dataset. These networks were also trained from scratch and only the best among the fine-tuned and the custom network of each model was selected for evaluation. The VOTCSW was not used with ResNet50V2, InceptionV3 and Xception mainly because they are 2D models and due to the fact that fine-tuning is only possible on images of size $224\times224$. Every model was trained with the Adam optimizer \cite{adam}, a batch size of $128$ and a learning rate of $10^{-3}$ for $100$ epochs.
\subsection{Results and discussion}
\begin{figure}[!htpb]
\centering
\subcaptionbox{CNNs accuracy over time.}{\includegraphics[scale=0.046]{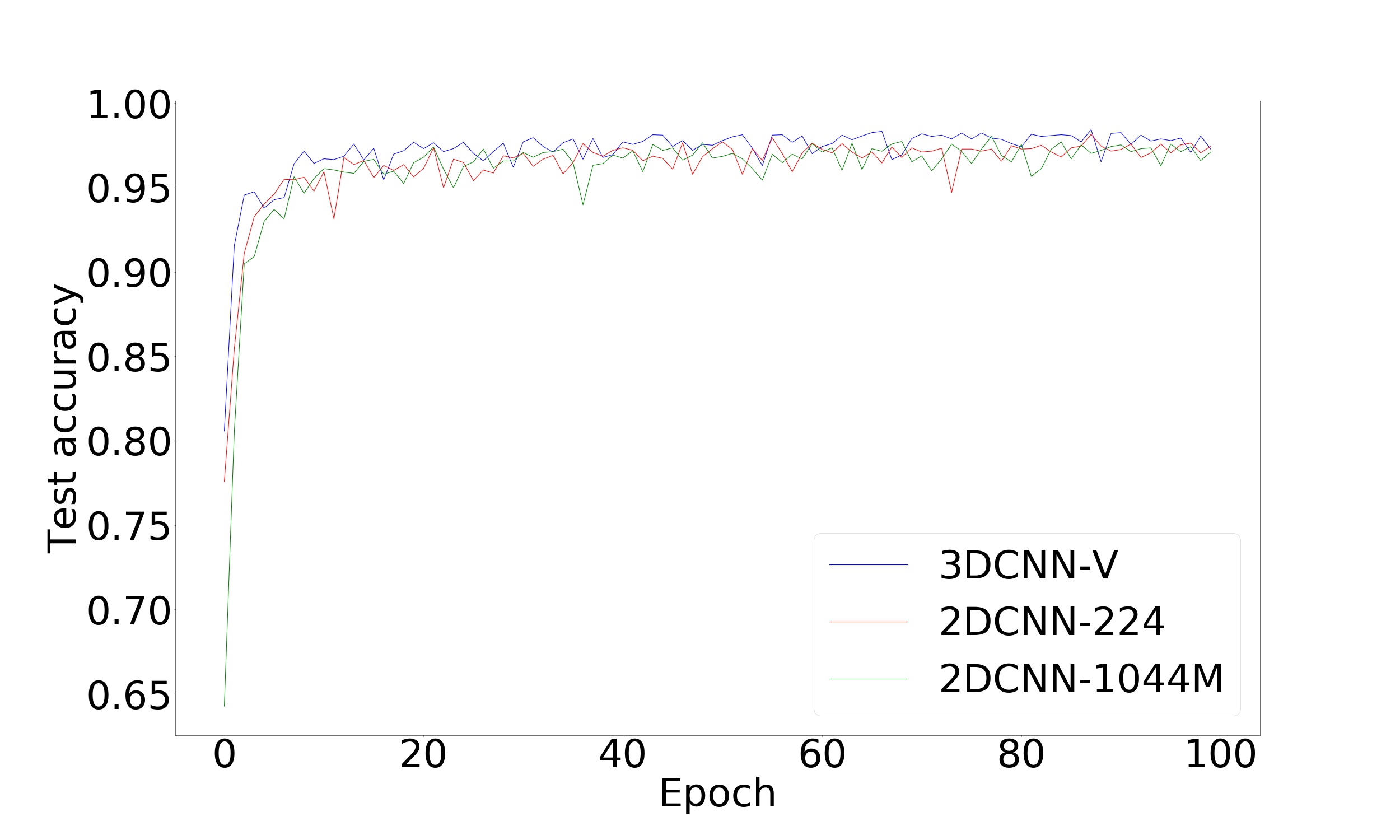}}\hspace{5pt}\subcaptionbox{NDPNNs accuracy over time.}{\includegraphics[scale=0.046]{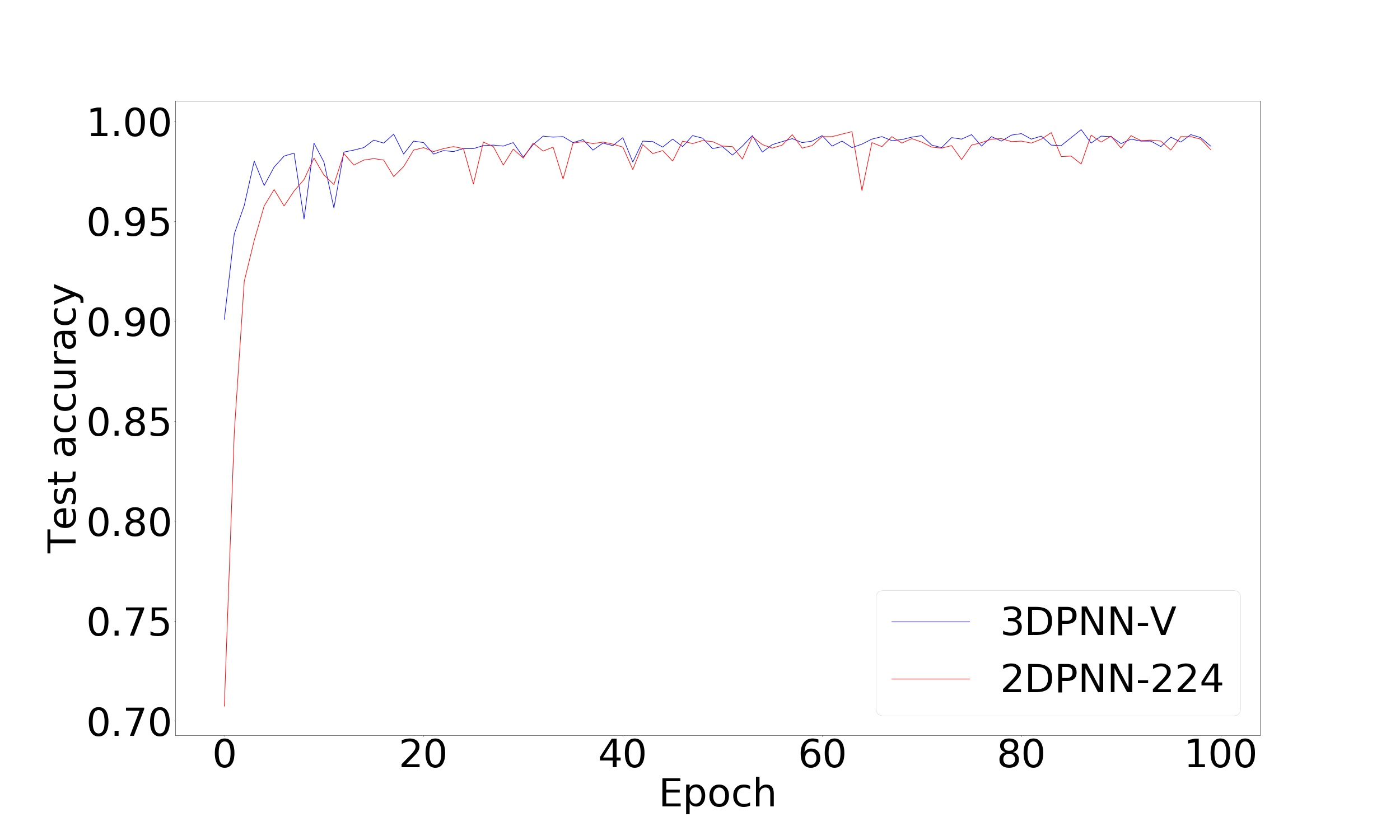}}
\vspace{10pt}
\caption{Evolution of the networks' test accuracies per epoch.}
\label{fig:accuracy_time}
\end{figure}

\begin{figure}[!htpb]
\centering
\subcaptionbox{Empty image.}{\includegraphics[scale=0.15]{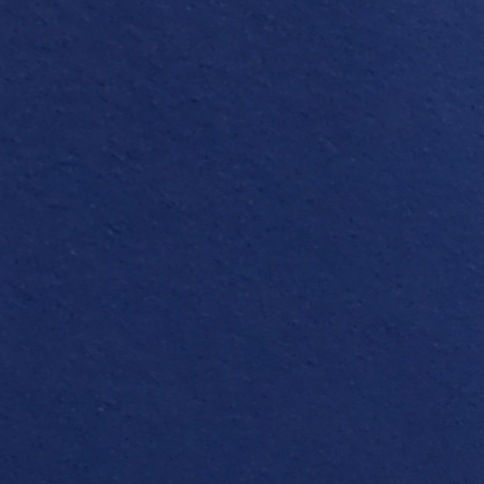}}\hspace{5pt}\subcaptionbox{Multiple plants in one image.}{\includegraphics[scale=0.06]{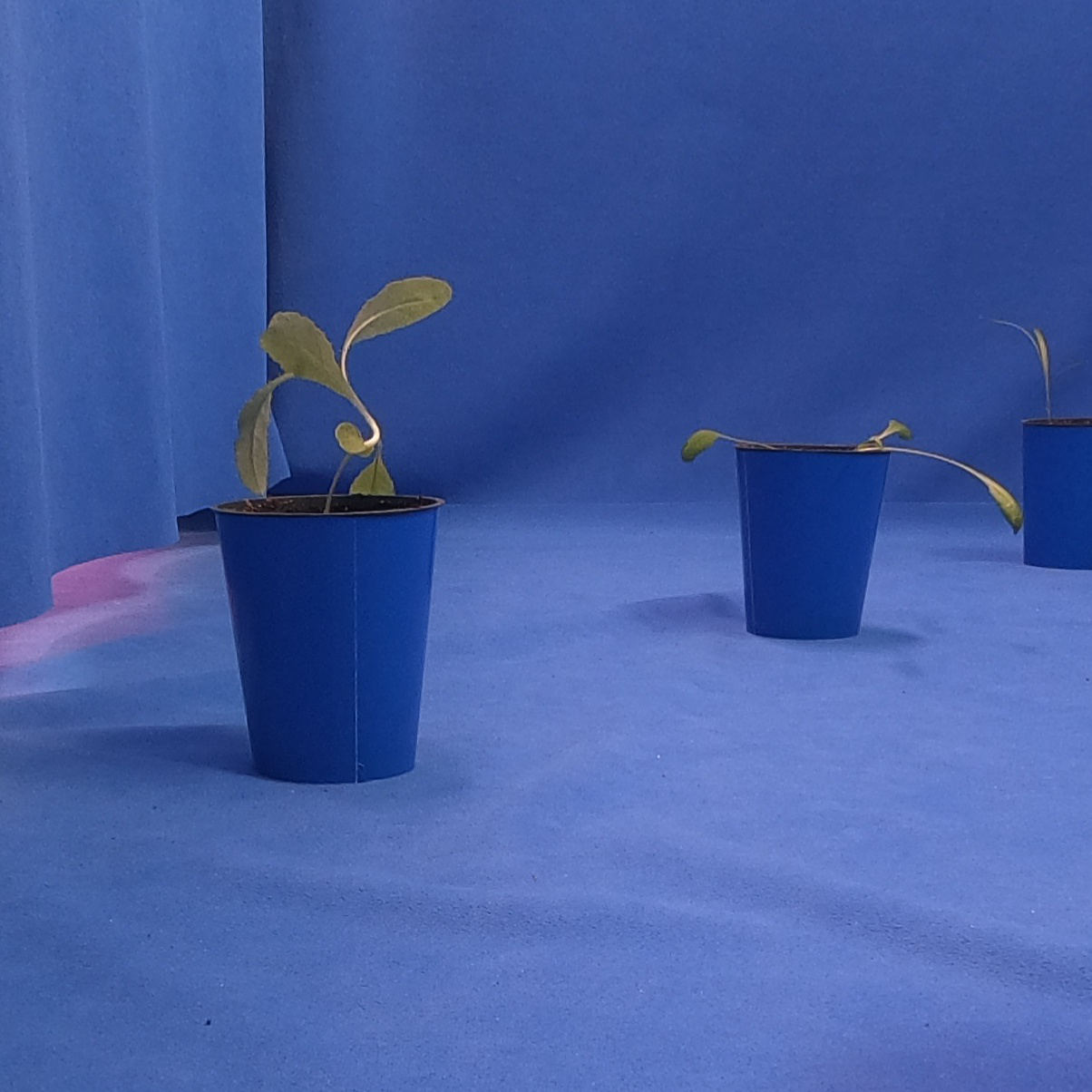}}
\vspace{10pt}
\caption{Examples of images wrongly classified by the 3DPNN-V model.}
\label{fig:wrongly_classified}
\end{figure}

\begin{figure}[!htpb]
\centering
\subcaptionbox{Regular 2D convolution kernel.}{\includegraphics[scale=0.06]{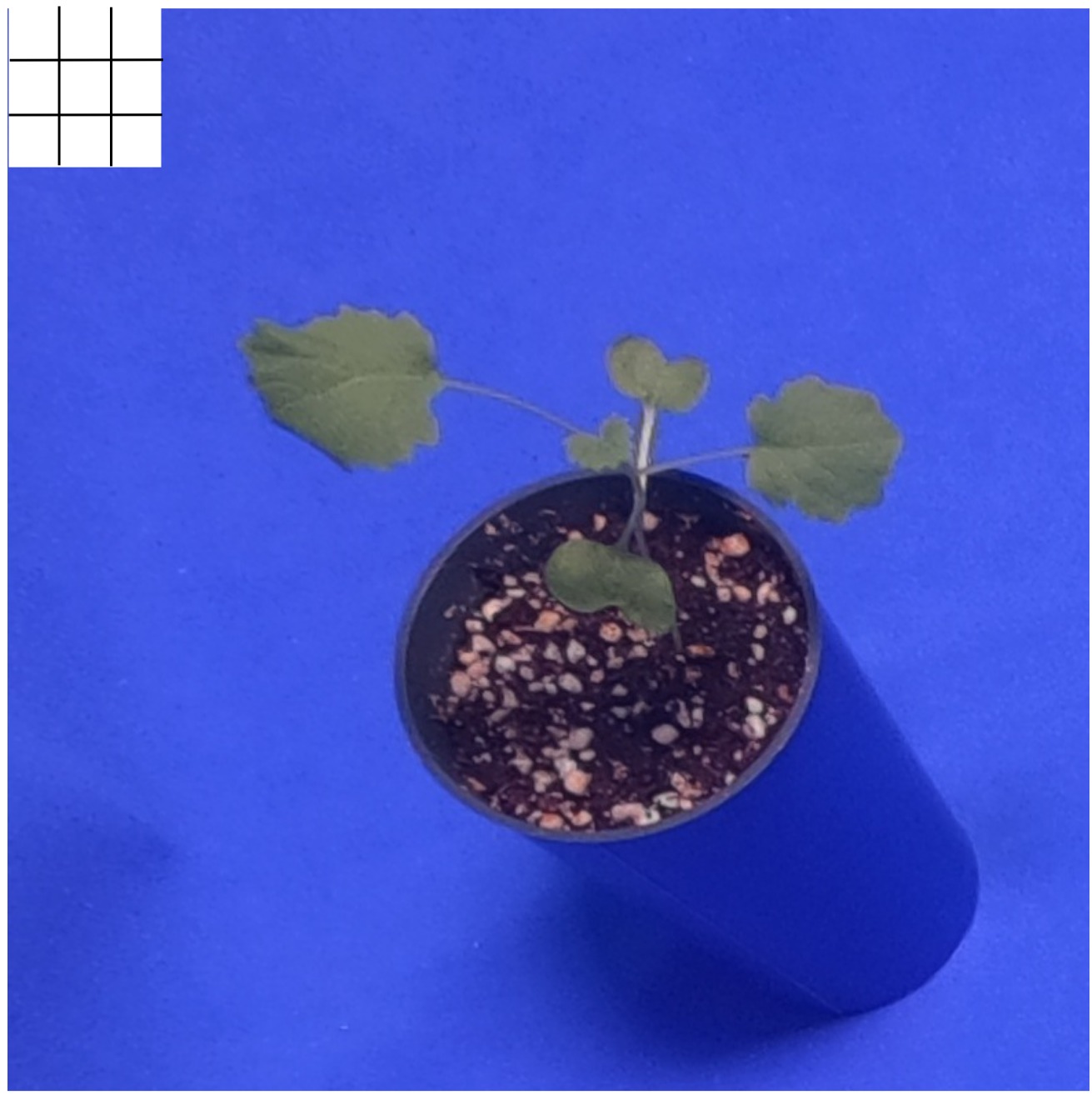}}\hspace{5pt}\subcaptionbox{VOTCSW 3D convolution kernel.}{\includegraphics[scale=0.06]{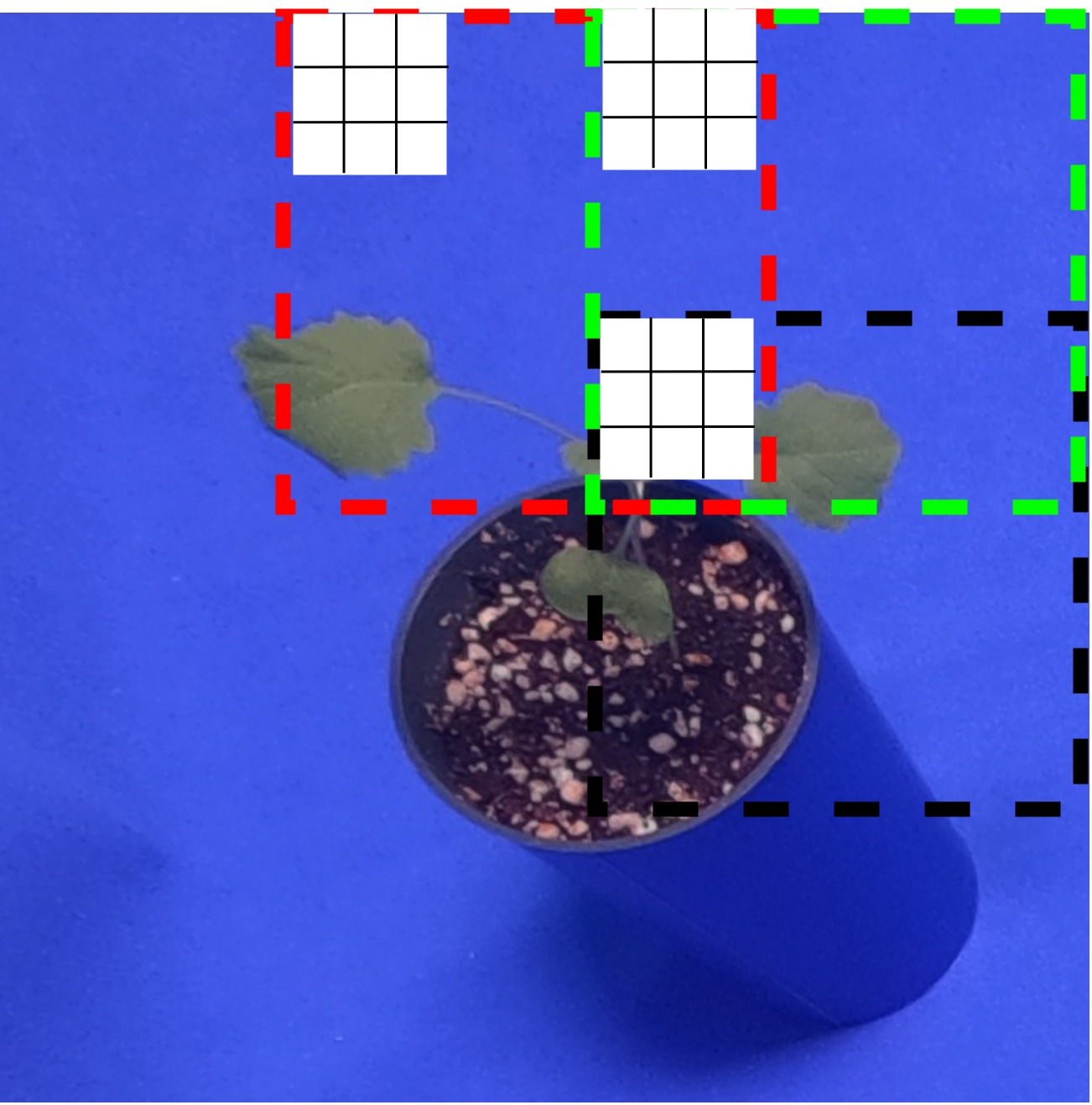}}
\vspace{10pt}
\caption{Comparison between a regular 2D convolution kernel and VOTCSW 3D convolution kernel.}
\label{fig:comparison}
\end{figure}

The grid search determined that, for all the networks, $64$ neurons in every layer (except the first) produces the best average results. The number of neurons for the 3DCNN layers was therefore determined to be $53$ according to Eq. (\ref{eq:N}) with $N_{1044}=257408$ which corresponds to the number of parameters of the feature extractor of the 2DCNN--1044P and the 2DCNN--1044M. Furthermore, ResNet50V2, InceptionV3 and Xception failed to produce decent results when they were trained from scratch. Therefore, only the fine-tuned networks were considered. Table \ref{table:performance} shows the best test accuracy, aggregated precision, aggregated recall, aggregated F1 score, the average inference time per sample and the number of trainable parameters of every model described above. The experiments show that the best 2DCNN model is the one trained on WSISCMC--224, and the best 3DCNN is the one trained with the vertical sliding pattern. Therefore, they were chosen to be extended to NDPNNs and the 2DPNN achieved a $99.48\%$ accuracy while the 3DPNN achieved a state-of-the-art $99.58\%$ accuracy. Their execution times and the number of trainable parameters were measured after the polynomial degree reduction described in Algorithm \ref{alg:degree_reduction} which determined that the first two layers of the 2DPNN could be reduced to a degree of $7$ and $2$ respectively while the remaining ones could be reduced to $1$ which represents $4.67$ times less parameters, and that the first three layers of the 3DPNN could be reduced to $6$, $2$ and $2$ respectively, while the remaining ones could be reduced to $1$ which represents $4.01$ times less parameters. The results also show that ResNet50V2, InceptionV3 and Xception failed to match the performance of the 2DPNN and the 3DPNN despite having more than 30 times the number of parameters. Furthermore, even though the accuracy of the 3DPNN is unmatched, the data generated by the VOTCSW method came with an increase in the spatio-temporal complexity of the model as it runs $3.28$ times slower than its 2D counterpart, and has $1.75$ times more parameters. However, both 2DCNN--1044P and 2DCNN--1044M did not provide satisfactory results compared to the 3DCNN models that run faster and have less parameters, which tends to show that the 3D representation created with the VOTCSW method is better than padding, magnifying and shrinking. Moreover, the 3DCNN models achieve better performance overall than the 2DCNN--224 model but they have more parameters and run slower than the 2DCNN--224 model which suggests that the VOTCSW method comes with the cost of slightly heavier but better models.
\newline\indent
\begin{table}[!htpb]
\resizebox{\columnwidth}{!}{
\begin{tabular}{lllllll}
& Performance\\\cline{2-7}
Model&Accuracy&Precision&Recall&F1 Score& Inference time (ms)&Parameters\\\hline
2DCNN--224 & 98.15&98.2&98.08&98.14&4.7&241,992\\\hline
2DCNN--1044P& 97.8&97.83&97.78&97.8&14.49&332,296\\\hline
2DCNN--1044M& 98.03&97.96&98.11&98.03&14.73&332,296\\\hline
3DCNN--H& 98.28&98.37&98.16&98.26&14.26&331,075\\\hline
3DCNN--V & 98.43&98.64&98.3&98.46&14.15&331,075\\\hline
3DCNN--S & 98.28&98.37&98.16&98.26&14.31&331,075\\\hline
2DPNN--224 & 99.48&99.53&99.33&99.42&4.98&265,608\\\hline
\textbf{3DPNN--V} & \textbf{99.58}&\textbf{99.69}&\textbf{99.36}&\textbf{99.52}&16.34&465,670\\\hline
ResNet50V2 & 98.25&98.17&98.28&98.22&28&23,581,192\\\hline
InceptionV3 & 97.78&97.64&97.83&97.73&38&21,819,176\\\hline
Xception & 97.9&97.85&98.03&97.93&31&21,819,176\\\hline

\end{tabular}
}
\caption{Accuracy, precision, recall, F1 score, inference time and number of trainable parameters of all the models trained on the reworked WSISCMC dataset. Bold values represent the highest value in their respective column.}
\label{table:performance}
\end{table}
An analysis of the generalization behavior of the networks shows that the 3DCNN--V model learns to generalize faster than the equally complex 2DCNN--1044M model and the 2DCNN--224 model as represented in Figure \ref{fig:accuracy_time}.(a). Furthermore, a more stable convergence is observed for the 3DCNN--V model which may be explained by the fact that the oversampling inherent to the VOTCSW method (refer to Proposition \ref{prop:oversampling}) has a regularization effect that smoothes the weight updates and enables a steadier training with a potential reduction of overfitting. These effects are also observed in the convergence of the NDPNN models represented in Figure \ref{fig:accuracy_time}.(b) where there is a clear gap between the convergence speed and stability of the 3DPNN--V and that of the 2DPNN--224.
A further analysis of the performance of the 3DPNN--V model shows that it was able to encapsulate enough information to perfectly recognize the least represented class in the dataset which is ``Wild Oat" as shown in Table \ref{table:confusion} outlining the confusion matrix of the 3DPNN--V model. Since only 17 images were wrongly classified, an in-depth investigation was performed. This investigation revealed that 10 images were showing a blue background as illustrated in Figure \ref{fig:wrongly_classified}.(a) and that 3 images were containing multiple plants in one image as shown in Figure \ref{fig:wrongly_classified}.(b). Moreover, upon further investigation, it was determined that the 3DPNN--V model correctly recognizes one of the plants present in all of the 3 multiple-plant images. Therefore, we can consider that the model is only wrong on 4 images since the problematic images contradict the task of single plant classification by either showing no plant or multiple ones. As a result, the 3DPNN--V model achieved an effective accuracy of $\dfrac{4004-17}{4004-17+4}=99.9\%$ when we removed the aberrant samples from the test set.
\begin{table}[!htpb]
\resizebox{\columnwidth}{!}{
\begin{tabular}{lllllllll}
Actual--Predicted&Canola&Dandelion&Canada Thistle&Wild Oat&Wild Buckwheat&Smartweed&Barnyard Grass&Yellow Foxtail\\\hline
Canola&752&1&1&0&1&0&0&0\\\hline
Dandelion&0&540&0&0&0&0&0&0\\\hline
Canada Thistle&0&0&545&0&0&0&0&0\\\hline
Wild Oat&0&0&0&126&0&0&0&0\\\hline
Wild Buckwheat&0&1&0&0&488&0&1&0\\\hline
Smartweed&0&1&1&0&0&145&2&0\\\hline
Barnyard Grass&1&0&0&0&0&0&939&0\\\hline
Yellow Foxtail&0&1&1&0&0&0&5&452\\\hline
\end{tabular}
}
\caption{Confusion matrix of the 3DPNN--V model.}
\label{table:confusion}
\end{table}
\newline\indent
We now investigate the reason why the VOTCSW method enables the creation of a model that generalizes better than one trained on resized images aside from its regularization-like behavior. The intuition behind the improved generalization comes from the fact that the VOTCSW method enables the model using a 3D convolution kernel to have a larger effective field of view than a 2D convolution kernel as illustrated in Figure \ref{fig:comparison} where the white squares represents a $3\times3$ convolution kernels, and the red, green and black dashed squares represent 3 consecutive overlapping windows generated by the VOTCSW method. The VOTCSW convolution kernel has three times more parameters and is more spatially dilated which enables it to take into account three distinct informative areas that may be distant such as leaves. Hence, this helps in creating ``spatially aware" models that can, not only achieve what regular 2D convolution models already do, but also establish a map of more complex spatial features.
\section{Conclusion}\label{chapter=plants:section=conclusion}
In this work, we used 2DPNNs and 3DPNNs to tackle the problem of plant species recognition on the WSISCMC dataset which contains plant images with variable size. We showed that the equations governing the behavior of a 1DPNN's forward and backward propagation are invariant with respect to the dimension of the signal which allowed us to generalize the 1DPNN to an NDPNN that can process 2D and 3D signals. In addition, we used a recently proposed polynomial degree reduction formula to develop a heuristic algorithm that performs layer-wise degree reduction on a pre-trained NDPNN which makes it lighter and faster while preserving its performance on the dataset it was trained on. We also formally developed the VOTCSW method which transforms any image from the dataset to a 3D representation with fixed size that is suitable for 3DCNNs and 3DPNNs. Furthermore, we designed a model development framework that makes use of the NDPNN layer-wise degree reduction heuristic and the VOTCSW to create a highly reliable NDPNN architecture for plant species recognition on the WSISCMC dataset. Moreover, we discovered that the current publicly available WSISCMC dataset can not be used with machine learning models without a mandatory preprocessing consisting in redistributing the samples of each class by occurrence and size to create a test set that has the same class distribution and size distribution as the training set. In addition, we evaluated the gain of using the VOTCSW method and we fine-tuned complex architectures such as ResNetV2, InceptionV3 and Xception. We also showed that the 3DPNN used in conjunction with the VOTCSW method and the NDPNN layer-wise degree reduction outperformed ResNetV2, InceptionV3 and Xception with far less spatio-temporal complexity as well as all other considered models and achieved a state-of-the-art 99.9\% accuracy on the WSISCMC dataset after determining the existence of aberrant samples in the dataset that contradict the fact that the dataset should only contain single-plant images. These results confirm the intuition proposed by Mehdipour Ghazi \textit{et al.} \cite{related:cnn_full_plant} who states that models with simpler architectures have the tendency to better learn from scratch than complex architectures. Besides, the experiments conducted show that the NDPNN layer-wise degree reduction heuristic is able to significantly compress a pre-trained NDPNN without altering its performance on the test set, which makes it a necessary postprocessing tool that has to be used in conjunction with NDPNNs. Furthermore, we also demonstrated that the VOTCSW method offers a better alternative than resizing when using the WSISCMC dataset which contains images with variable sizes and that the 3D representation it creates is more informative than a resized 2D representation. Although the WSISCMC dataset has some issues, we can safely declare that the EAGL--I system has the potential to produce highly relevant massive datasets, provided that the authors impose a stricter control on the data acquisition process and ensure that classes are balanced.
\newline\indent
Despite its effectiveness, the NDPNN layer-wise degree reduction heuristic was only applied after the training of an NDPNN was completed. This is the most time consuming process and the most risky -- in terms of stability -- as 1DPNNs were proven to show some instability when trained with unbounded activation functions and this instability is expected to be observed on 2DPNNs and 3DPNNs. The instability can potentially be reduced by lowering the degree of each layer of the model either before or during training (hence the potential to use this heuristic in the model validation process) instead of after a model is trained. Furthermore, the heuristic is based on determining the smallest symmetric interval that contains the values of a given layer's output regardless of the channel/neuron. This implies that some polynomials are over-reduced, meaning that they are reduced on a bigger interval than the one they produce their output from. A potential solution to this is to determine the smallest symmetric interval that contains the values of a given neuron's output to produce a finer and more accurate reduction. As for the VOTCSW method, although it enabled the creation of highly accurate models on the WSISCMC dataset, there is not enough evidence to claim that it can improve the results on any dataset that contains images with variable size. Additionally, the increase in model performance may not always justify the size and parameter overhead that it introduces compared to simply shrinking images for applications that are memory bound. Besides, there is no clear indication on how to determine the minimum ratio $\dfrac{\tilde{H}_{max}}{\tilde{H}_{min}}$ or the adequate parameters $M$, $\alpha_{min}$, $\alpha_{max}$ and $h$ that can maximize the performance of a model on a given dataset. 
\newline\indent
Therefore, future work will focus on applying the VOTCSW method on bigger and more varied datasets in order to determine if the performance improvement that it introduces to models has any statistical significance. We will also aim to determine how the choice of the VOTCSW parameters influence the performance of the trained models, with an emphasis on creating a more dataset-specific set of rules for the method's use. We also plan on exploiting the NDPNN layer-wise degree reduction heuristic for validating models before training, which can be a safer alternative to the one it was used in this work. In addition, we intend to apply the polynomial reduction on the neuronal level in order to obtain more stable and accurate polynomials than the ones obtained with the layer-wise reduction, which may enable the degree of a layer to be further reduced.
\section*{Acknowledgement}
This work was funded by the Mitacs Globalink Graduate Fellowship Program (no. FR41237) and the NSERC Discovery Grants Program (nos. 194376, 418413). The authors wish to thank The University of Winnipeg’s Faculty of Graduate Studies for the UW Graduate Studies Scholarship, and Dr. Michael Beck for his valuable assistance with the WSISCMC dataset.
\begin{appendices}
\section{Mathematical Proofs}
\subsubsection*{Proof of Proposition \ref{prop:alpha}}
\begin{proof}[]
From Eq.(\ref{eq:M}), we have
\begin{flalign*}
\begin{split}
&M=\left(\dfrac{H-h}{(1-\alpha)h}+1\right)\left(\dfrac{W-w}{(1-\alpha)w}+1\right)\\
&\iff hwM(1-\alpha)^2=\left(H-h+h(1-\alpha)\right)\left(W-w+w(1-\alpha)\right)\\
&\iff hwM(1-\alpha)^2=(H-h)(W-w)+((H-h)w+(W-w)h)(1-\alpha)+hw(1-\alpha)^2\\
&\iff hw(M-1)(1-\alpha)^2-((H-h)w+(W-w)h)(1-\alpha)-(H-h)(W-w)=0.
\end{split} &&
\end{flalign*}
This is a second degree equation with $(1-\alpha)$ as unknown. Therefore, the discriminant is
\begin{flalign*}
\Delta=((H-h)w+(W-w)h)^2+4hw(H-h)(W-w)(M-1).&&
\end{flalign*}
Since $M$ is the number of windows, it is necessarily greater than 1. Moreover, $H>h$ and $W>w$ because the window size is always smaller than the image size. Consequently $\Delta>0$ and we have two solutions expressed as such:
\begin{flalign*}
1-\alpha_{1,2}=\dfrac{((H-h)w+(W-w)h)\pm\sqrt{\Delta}}{2hw(M-1)}.&&
\end{flalign*}
However, $\sqrt{\Delta}> ((H-h)w+(W-w)h)$ and since $\alpha\in[0,1[$ by definition, then $1-\alpha>0$ and only the following valid solution remains:
\begin{flalign*}
\alpha=1-\dfrac{((H-h)w+(W-w)h)+\sqrt{\Delta}}{2hw(M-1)}.&&
\end{flalign*}
\end{proof}
\subsubsection*{Proof of Proposition \ref{prop:alpha_positive}}
\begin{proof}[]
We suppose that $\alpha\geq0$.
\begin{flalign*}
\begin{split}
&\alpha\geq0\iff 1-\alpha\leq1\iff\dfrac{((H-h)w+(W-w)h)+\sqrt{\Delta}}{2hw(M-1)}\leq1\\
&\underset{M>1}{\iff}\sqrt{\Delta}\leq 2hw(M-1) - ((H-h)w+(W-w)h)\\
&\iff\\
&((H-h)w+(W-w)h)^2+4hw(H-h)(W-w)(M-1)\\
&\leq 4h^2w^2(M-1)^2 + ((H-h)w+(W-w)h)^2 - 4hw((H-h)w+(W-w)h)(M-1)\\
\end{split}
&
\end{flalign*}
\begin{flalign*}
\begin{split}
&\iff 4hw(M-1)\left((H-h)(W-w)+(H-h)w+(W-w)h-hw(M-1)\right)\leq0\\
&\underset{M>1}{\iff}\left((H-h)(W-w)+(H-h)w+(W-w)h-hw(M-1)\right)\leq0\\
&\iff HW - Hw - hW +hw + Hw - hw + hW - hw - hwM + hw  \leq0\\
&\iff HW - hwM \leq0\\
&\iff hwM \geq HW
\end{split}&&
\end{flalign*}
\end{proof}
\subsubsection*{Proof of Proposition \ref{prop:alpha_simple}}
\begin{proof}[]
By combining Eq. (\ref{eq:H&W}) and Eq. (\ref{eq:M_N_h}), we have
\begin{flalign*}
\begin{split}
M = (N_h+1)^2=\left(\dfrac{H-h}{(1-\alpha)h}+1\right)^2&\iff(1-\alpha)^2h^2M=(H-\alpha h)^2\\
&\iff(1-\alpha)h\sqrt{M}=H-\alpha h\\
&\iff\alpha=\dfrac{\sqrt{M}h-H}{h(\sqrt{M}-1)}.
\end{split}&&
\end{flalign*}
The positivity of $\alpha$ is verified when $h\geq \dfrac{H}{\sqrt{M}}$.
\end{proof}
\subsubsection*{Proof of Proposition \ref{prop:h}}
\begin{proof}[]
We know that $H_{min}\leq H\leq H_{max}$. Consequently, by using Eq. (\ref{eq:alpha_simple}) we have
\begin{flalign}\label{eq:alpha_min_max}
\alpha_{min}\leq\dfrac{\sqrt{M}h-H_{max}}{h(\sqrt{M}-1)}\leq \alpha\leq\dfrac{\sqrt{M}h-H_{min}}{h(\sqrt{M}-1)}\leq\alpha_{max}.&& 
\end{flalign}
Therefore, by extracting $h$ from Eq. (\ref{eq:alpha_min_max}) we obtain
\begin{flalign*}
\dfrac{H_{max}}{\sqrt{M}-\alpha_{min}(\sqrt{M}-1)}\leq h\leq \dfrac{H_{min}}{\sqrt{M}-\alpha_{max}(\sqrt{M}-1)}.&&
\end{flalign*}
\end{proof}
\subsubsection*{Proof of Theorem \ref{theorem:condition on the parameters}}
\begin{proof}[]
The proof of this theorem is based solely on the result of Proposition \ref{prop:h}. The condition for this Proposition to be valid is
\begin{flalign*}
\dfrac{H_{max}}{\sqrt{M}-\alpha_{min}(\sqrt{M}-1)}\leq \dfrac{H_{min}}{\sqrt{M}-\alpha_{max}(\sqrt{M}-1)}&&.
\end{flalign*}
Therefore we get
\begin{flalign}\label{eq:root_eq_theorem}
H_{max}\left(\sqrt{M}-\alpha_{max}(\sqrt{M}-1)\right)\leq H_{min}\left(\sqrt{M}-\alpha_{min}(\sqrt{M}-1)\right).&&
\end{flalign}
This equation establishes a relationship between $M$, $\alpha_{min}$ and $\alpha_{max}$ and the conditions stated in the theorem are all derived from it. We will only prove the case where $\alpha_{max}$ is determined first, then $\alpha_{min}$ then $M$ because it illustrates how the 5 other cases are proved. We begin by isolating $\sqrt{M}$ from Eq. (\ref{eq:root_eq_theorem}) as such:
\begin{flalign*}
\sqrt{M}\left(H_{max}(1-\alpha_{max})-H_{min}(1-\alpha_{min})\right)\leq H_{min}\alpha_{min}-H_{max}\alpha_{max}.&&
\end{flalign*}
The right term of the inequality is negative since $H_{min}\leq H_{max}$ and $\alpha_{min}\leq\alpha_{max}$ by definition. Consequently, if the term $\left(H_{max}(1-\alpha_{max})-H_{min}(1-\alpha_{min})\right)$ was positive, it would result in $\sqrt{M}\leq0$ which is false by definition. Therefore, it must be negative for $M$ to exist. As a result, we obtain the following condition:
\begin{flalign*}
\sqrt{M}&\geq\dfrac{H_{max}\alpha_{max}-H_{min}\alpha_{min}}{H_{min}(1-\alpha_{min})-H_{max}(1-\alpha_{max})}.&&
\end{flalign*}
The fact that $\left(H_{max}(1-\alpha_{max})-H_{min}(1-\alpha_{min})\right)\leq0$ leads to
\begin{flalign*}
\alpha_{min}\leq 1-\dfrac{H_{max}}{H_{min}}(1-\alpha_{max}).&&
\end{flalign*}
However, $\alpha_{min}\geq0$ by definition, so the right term of the inequality has to be positive for $\alpha_{min}$ to exist. Thus, we obtain the following condition:
\begin{flalign*}
\alpha_{max}\geq1-\dfrac{H_{min}}{H_{max}}.&&
\end{flalign*}
The other cases are proved by using the same reasoning.
\end{proof}
\subsubsection*{Proof of Proposition \ref{prop:oversampling}}
\begin{proof}[]
We designate by $(x,y)$ a pixel in a given image. We suppose that $h$, $M$, and $\gamma$ were chosen prior to the use of the VOTCSW method and that the overlap $\alpha$ was calculated for the image. For the pixel $(x,y)$ to be contained in a given window $(n,m)$ such that $(n,m)\in[\![0,\sqrt{M}-1]\!]^2$, the following conditions need to be verified:
\begin{flalign*}
\begin{cases}
(1-\alpha)hn&\leq x\leq (1-\alpha)hn+h\\
(1-\alpha)\gamma hm&\leq y\leq (1-\alpha)\gamma hm+\gamma h\\
\end{cases}.&&
\end{flalign*} 
By extracting $n$ and $m$ from the previous conditions, we obtain
\begin{flalign*}
\begin{cases}
\dfrac{x}{(1-\alpha)h}-\dfrac{1}{1-\alpha}&\leq n \leq \dfrac{x}{(1-\alpha)h}\\
\dfrac{y}{(1-\alpha)\gamma h}-\dfrac{1}{1-\alpha}&\leq m \leq \dfrac{y}{(1-\alpha)\gamma h}		
\end{cases},&&
\end{flalign*}
which determines what values of $n$ and $m$ are valid for a window to contain the pixel $(x,y)$. Since $n$ is an integer, it can take at most $\dfrac{x}{(1-\alpha)h}-\left(\dfrac{x}{(1-\alpha)h}-\dfrac{1}{1-\alpha}\right)=\dfrac{1}{1-\alpha}$ different values in $[\![0,\sqrt{M}-1]\!]$, and the same can be inferred for $m$. Since every combination of $n$ and $m$ that follows the previous conditions can determine a window that contains the pixel $(x,y)$, the total number of times the pixel $(x,y)$ is present in a window is at most $\dfrac{1}{(1-\alpha)^2}$.
\end{proof}
\end{appendices}
\bibliography{main}

\begin{thebibliography}{10}

\bibitem{intro:multi_disciplinary}
C.~R. Stark, ``Adopting multidisciplinary approaches to sustainable agriculture
  research: Potentials and pitfalls,'' {\em Am J Altern Agric}, vol.~10, no.~4,
  p.~180–183, 1995.

\bibitem{intro:multi_disciplinary_1}
M.~Barbercheck, N.~E. Kiernan, A.~G. Hulting, S.~Duiker, J.~Hyde, H.~Karsten,
  and E.~Sanchez, ``Meeting the ‘multi-’ requirements in organic
  agriculture research: Successes, challenges and recommendations for
  multifunctional, multidisciplinary, participatory projects,'' {\em Renewable
  Agric Food Syst}, vol.~27, no.~2, p.~93–106, 2012.

\bibitem{intro:multi_disciplinary_2}
A.~Luca, G.~Molari, G.~Seddaiu, A.~Toscano, G.~Bombino, L.~Ledda, M.~Milani,
  and M.~Vittuari, ``Multidisciplinary and innovative methodologies for
  sustainable management in agricultural systems,'' {\em Environ Eng Manage J},
  vol.~14, no.~7, pp.~1571--1581, 2015.

\bibitem{intro:deep_knowledge}
M.~J. Paul, A.~Watson, and C.~A. Griffiths, ``{Linking fundamental science to
  crop improvement through understanding source and sink traits and their
  integration for yield enhancement},'' {\em J Exp Bot}, vol.~71,
  pp.~2270--2280, 10 2019.

\bibitem{intro:deep_knowledge_1}
N.~K. Fageria, V.~C. Baligar, and Y.~C. Li, ``The role of nutrient efficient
  plants in improving crop yields in the twenty first century,'' {\em J Plant
  Nutr}, vol.~31, no.~6, pp.~1121--1157, 2008.

\bibitem{intro:deep_knowledge_2}
N.~Senapati, H.~E. Brown, and M.~A. Semenov, ``Raising genetic yield potential
  in high productive countries: Designing wheat ideotypes under climate
  change,'' {\em Agric For Meteorol}, vol.~271, pp.~33--45, 2019.

\bibitem{intro:machinery}
L.~F.~P. Oliveira, A.~P. Moreira, and M.~F. Silva, ``Advances in agriculture
  robotics: A state-of-the-art review and challenges ahead,'' {\em Robotics},
  vol.~10, no.~2, 2021.

\bibitem{intro:machinery_1}
T.~Duckett, S.~Pearson, S.~Blackmore, and B.~Grieve, ``Agricultural robotics:
  The future of robotic agriculture,'' {\em CoRR}, vol.~abs/1806.06762, 2018.

\bibitem{intro:machinery_2}
J.~Relf-Eckstein, A.~T. Ballantyne, and P.~W. Phillips, ``Farming reimagined: A
  case study of autonomous farm equipment and creating an innovation
  opportunity space for broadacre smart farming,'' {\em NJAS Wageningen J Life
  Sci}, vol.~90-91, p.~100307, 2019.

\bibitem{intro:emergence}
K.~Jha, A.~Doshi, P.~Patel, and M.~Shah, ``A comprehensive review on automation
  in agriculture using artificial intelligence,'' {\em Artif Intell Agric},
  vol.~2, pp.~1--12, 2019.

\bibitem{intro:agriculture_4}
D.~C. Rose and J.~Chilvers, ``Agriculture 4.0: Broadening responsible
  innovation in an era of smart farming,'' {\em Front Sustainable Food Syst},
  vol.~2, 2018.

\bibitem{intro:digital_agriculture}
A.~G. Green, A.-R. Abdulai, E.~Duncan, A.~Glaros, M.~Campbell, R.~Newell,
  P.~Quarshie, K.~B. KC, L.~Newman, E.~Nost, and E.~D.~G. Fraser, ``A scoping
  review of the digital agricultural revolution and ecosystem services:
  implications for canadian policy and research agendas,'' {\em FACETS},
  vol.~6, pp.~1955--1985, 2021.

\bibitem{intro:precision_agriculture}
I.~Cisternas, I.~Velásquez, A.~Caro, and A.~Rodríguez, ``Systematic
  literature review of implementations of precision agriculture,'' {\em Comput
  Electron Agric}, vol.~176, p.~105626, 2020.

\bibitem{intro:planting}
N.~Srinivasan, P.~Prabhu, S.~S. Smruthi, N.~V. Sivaraman, S.~J. Gladwin,
  R.~Rajavel, and A.~R. Natarajan, ``Design of an autonomous seed planting
  robot,'' in {\em IEEE Reg 10 Humanit Technol Conf.}, pp.~1--4, 2016.

\bibitem{intro:planting_1}
S.~Sukkarieh, ``Mobile on-farm digital technology for smallholder farmers,''
  {\em Transforming Lives and Livelihoods: The Digital Revolution in
  Agriculture}, no.~2059-2018-203, p.~9, 2017.

\bibitem{intro:planting_2}
M.~U. Hassan, M.~Ullah, and J.~Iqbal, ``Towards autonomy in agriculture: Design
  and prototyping of a robotic vehicle with seed selector,'' in {\em Int Conf
  Rob Artif Intel}, pp.~37--44, 2016.

\bibitem{intro:harvesting}
S.~Birrell, J.~Hughes, J.~Y. Cai, and F.~Iida, ``A field-tested robotic
  harvesting system for iceberg lettuce,'' {\em J Field Rob}, vol.~37, no.~2,
  pp.~225--245, 2020.

\bibitem{intro:harvesting_1}
Y.~Ge, Y.~Xiong, G.~L. Tenorio, and P.~J. From, ``Fruit localization and
  environment perception for strawberry harvesting robots,'' {\em IEEE Access},
  vol.~7, pp.~147642--147652, 2019.

\bibitem{intro:harvesting_2}
D.~SepúLveda, R.~Fernández, E.~Navas, M.~Armada, and P.~González-De-Santos,
  ``Robotic aubergine harvesting using dual-arm manipulation,'' {\em IEEE
  Access}, vol.~8, pp.~121889--121904, 2020.

\bibitem{intro:machine_learning}
V.~Meshram, K.~Patil, V.~Meshram, D.~Hanchate, and S.~Ramkteke, ``Machine
  learning in agriculture domain: A state-of-art survey,'' {\em Artif Intell
  Life Sci}, vol.~1, p.~100010, 2021.

\bibitem{intro:efforts}
D.~I. Patrício and R.~Rieder, ``Computer vision and artificial intelligence in
  precision agriculture for grain crops: A systematic review,'' {\em Comput
  Electron Agric}, vol.~153, pp.~69--81, 2018.

\bibitem{intro:efforts_1}
A.~Kamilaris and F.~X. Prenafeta-Boldú, ``Deep learning in agriculture: A
  survey,'' {\em Comput Electron Agric}, vol.~147, pp.~70--90, 2018.

\bibitem{intro:disease}
R.~Sujatha, J.~M. Chatterjee, N.~Jhanjhi, and S.~N. Brohi, ``Performance of
  deep learning vs machine learning in plant leaf disease detection,'' {\em
  Microprocess Microsyst}, vol.~80, p.~103615, 2021.

\bibitem{intro:growth}
D.~Sivakumar, K.~SuriyaKrishnaan, P.~Akshaya, G.~Anuja, and G.~Devadharshini,
  ``Computerized growth analysis of seeds using deep learning method,'' {\em
  Int J of Recent Technol Eng}, vol.~7, no.~6S5, 2019.

\bibitem{intro:irrigation}
A.~Morellos, X.-E. Pantazi, D.~Moshou, T.~Alexandridis, R.~Whetton,
  G.~Tziotzios, J.~Wiebensohn, R.~Bill, and A.~M. Mouazen, ``Machine learning
  based prediction of soil total nitrogen, organic carbon and moisture content
  by using vis-nir spectroscopy,'' {\em Biosyst Eng}, vol.~152, pp.~104--116,
  2016.
\newblock Proximal Soil Sensing – Sensing Soil Condition and Functions.

\bibitem{intro:feature_engineering}
J.~Wäldchen, M.~Rzanny, M.~Seeland, and P.~Mäder, ``Automated plant species
  identification—trends and future directions,'' {\em PLoS Comput Biol},
  vol.~14, pp.~1--19, 04 2018.

\bibitem{intro:feature_engineering_1}
T.~Jin, X.~Hou, P.~Li, and F.~Zhou, ``A novel method of automatic plant species
  identification using sparse representation of leaf tooth features,'' {\em
  PLOS ONE}, vol.~10, pp.~1--20, 10 2015.

\bibitem{intro:feature_engineering_2}
P.~Barré, B.~C. Stöver, K.~F. Müller, and V.~Steinhage, ``Leafnet: A
  computer vision system for automatic plant species identification,'' {\em
  Ecol Inf}, vol.~40, pp.~50--56, 2017.

\bibitem{intro:dataset}
N.~Kumar, P.~N. Belhumeur, A.~Biswas, D.~W. Jacobs, W.~J. Kress, I.~Lopez, and
  J.~V.~B. Soares, ``Leafsnap: A computer vision system for automatic plant
  species identification,'' in {\em Eur Conf Comput Vision}, October 2012.

\bibitem{intro:dataset_1}
A.~Olsen, D.~A. Konovalov, B.~Philippa, P.~Ridd, J.~C. Wood, J.~Johns,
  W.~Banks, B.~Girgenti, O.~Kenny, J.~Whinney, B.~Calvert, M.~R. Azghadi, and
  R.~D. White, ``Deepweeds: A multiclass weed species image dataset for deep
  learning,'' {\em Sci Rep}, vol.~9, p.~2058, Feb 2019.

\bibitem{intro:dataset_2}
{\em Plant Identification in an Open-world ({LifeCLEF} 2016)}, vol.~CEUR
  Workshop Proceedings of {\em CLEF: Conference and Labs of the Evaluation
  Forum}, ({\'E}vora, Portugal), Sept. 2016.

\bibitem{intro:EAGL-I}
M.~A. Beck, C.-Y. Liu, C.~P. Bidinosti, C.~J. Henry, C.~M. Godee, and
  M.~Ajmani, ``An embedded system for the automated generation of labeled plant
  images to enable machine learning applications in agriculture,'' {\em PLOS
  ONE}, vol.~15, pp.~1--23, 12 2020.

\bibitem{intro:EAGL-I_big_dataset}
M.~A. Beck, C.~Liu, C.~P. Bidinosti, C.~J. Henry, C.~M. Godee, and M.~Ajmani,
  ``Presenting an extensive lab- and field-image dataset of crops and weeds for
  computer vision tasks in agriculture,'' {\em CoRR}, vol.~abs/2108.05789,
  2021.

\bibitem{intro:WSISCMC}
M.~A. Beck, C.-Y. Liu, C.~P. Bidinosti, C.~J. Henry, C.~M. Godee, and
  M.~Ajmani, ``Weed seedling images of species common to manitoba, canada,''
  2021.
\newblock \url{https://doi.org/10.5061/dryad.gtht76hhz}.

\bibitem{intro:translation}
T.~Park, A.~A. Efros, R.~Zhang, and J.~Zhu, ``Contrastive learning for unpaired
  image-to-image translation,'' {\em CoRR}, vol.~abs/2007.15651, 2020.

\bibitem{intro:1DPNN}
H.~{Ben Abdallah}, C.~J. Henry, and S.~Ramanna, ``1-dimensional polynomial
  neural networks for audio signal related problems,'' {\em Knowledge-Based
  Syst}, vol.~240, p.~108174, 2022.

\bibitem{intro:reduction}
H.~{Ben Abdallah}, C.~J. Henry, and S.~Ramanna, ``Polynomial degree reduction
  in the l2-norm on a symmetric interval for the canonical basis,'' {\em
  Results Appl Math}, vol.~12, p.~100185, 2021.

\bibitem{intro:resnet}
K.~He, X.~Zhang, S.~Ren, and J.~Sun, ``Identity mappings in deep residual
  networks,'' in {\em Eur Conf Comput Vision} (B.~Leibe, J.~Matas, N.~Sebe, and
  M.~Welling, eds.), (Cham), pp.~630--645, Springer International Publishing,
  2016.

\bibitem{intro:inception}
C.~Szegedy, V.~Vanhoucke, S.~Ioffe, J.~Shlens, and Z.~Wojna, ``Rethinking the
  inception architecture for computer vision,'' in {\em IEEE Conf Comput Vision
  Pattern Recognit}, pp.~2818--2826, 2016.

\bibitem{intro:xception}
F.~Chollet, ``Xception: Deep learning with depthwise separable convolutions,''
  in {\em IEEE Conf Comput Vision Pattern Recognit}, pp.~1800--1807, 2017.

\bibitem{related:leaves}
S.~Zhang, W.~Huang, Y.~an~Huang, and C.~Zhang, ``Plant species recognition
  methods using leaf image: Overview,'' {\em Neurocomputing}, vol.~408,
  pp.~246--272, 2020.

\bibitem{related:dataset}
S.~G. Wu, F.~S. Bao, E.~Y. Xu, Y.-X. Wang, Y.-F. Chang, and Q.-L. Xiang, ``A
  leaf recognition algorithm for plant classification using probabilistic
  neural network,'' in {\em IEEE Int Symp Signal Process Inf Technol},
  pp.~11--16, 2007.

\bibitem{related:feature_engineering}
S.~Purohit, R.~Viroja, S.~Gandhi, and N.~Chaudhary, ``Automatic plant species
  recognition technique using machine learning approaches,'' in {\em Int Conf
  Comput Network Commun}, pp.~710--719, 2015.

\bibitem{related:leaf_deep_learning}
X.~Wang, C.~Zhang, and S.~Zhang, ``Multiscale convolutional neural networks
  with attention for plant species recognition,'' {\em Comput Intell Neurosci},
  vol.~2021, pp.~5529905--5529905, Jul 2021.

\bibitem{related:cnn_full_plant}
M.~{Mehdipour Ghazi}, B.~Yanikoglu, and E.~Aptoula, ``Plant identification
  using deep neural networks via optimization of transfer learning
  parameters,'' {\em Neurocomputing}, vol.~235, pp.~228--235, 2017.

\bibitem{related:vgg_net}
K.~Simonyan and A.~Zisserman, ``Very deep convolutional networks for
  large-scale image recognition,'' {\em CoRR}, vol.~abs/1409.1556, 2015.

\bibitem{related:google_net}
C.~Szegedy, W.~Liu, Y.~Jia, P.~Sermanet, S.~E. Reed, D.~Anguelov, D.~Erhan,
  V.~Vanhoucke, and A.~Rabinovich, ``Going deeper with convolutions,'' {\em
  CoRR}, vol.~abs/1409.4842, 2014.

\bibitem{related:alex_net}
A.~Krizhevsky, I.~Sutskever, and G.~E. Hinton, ``Imagenet classification with
  deep convolutional neural networks,'' in {\em Adv Neural Inf Process Syst}
  (F.~Pereira, C.~J.~C. Burges, L.~Bottou, and K.~Q. Weinberger, eds.),
  vol.~25, Curran Associates, Inc., 2012.

\bibitem{related:poly_degree_2}
Y.~Wang, L.~Xie, C.~Liu, S.~Qiao, Y.~Zhang, W.~Zhang, Q.~Tian, and A.~Yuille,
  ``Sort: Second-order response transform for visual recognition,'' in {\em
  IEEE Int Conf Comput Vision}, pp.~1368--1377, 2017.

\bibitem{related:le_net}
Y.~Lecun, L.~Bottou, Y.~Bengio, and P.~Haffner, ``Gradient-based learning
  applied to document recognition,'' {\em Proc IEEE}, vol.~86, no.~11,
  pp.~2278--2324, 1998.

\bibitem{related:wr_net}
S.~Zagoruyko and N.~Komodakis, ``Wide residual networks,'' in {\em Proc. Br.
  Mach. Vision Conf.} (E.~R.~H. Richard C.~Wilson and W.~A.~P. Smith, eds.),
  pp.~87.1--87.12, BMVA Press, September 2016.

\bibitem{related:cifar}
A.~Krizhevsky and G.~Hinton, ``Learning multiple layers of features from tiny
  images,'' Tech. Rep.~0, University of Toronto, Toronto, Ontario, 2009.

\bibitem{related:svhn}
Y.~Netzer, T.~Wang, A.~Coates, A.~Bissacco, B.~Wu, and A.~Y. Ng, ``Reading
  digits in natural images with unsupervised feature learning,'' in {\em NIPS
  Workshop Deep Learn. Unsuperv. Feature Learn.}, 2011.

\bibitem{related:poly_degree_2_rnn}
T.~Hughes and K.~Mierle, ``Recurrent neural networks for voice activity
  detection,'' in {\em IEEE Int. Conf. Acoust. Speech Signal Process.},
  pp.~7378--7382, IEEE, 2013.

\bibitem{related:poly_degree_3}
F.~Babiloni, I.~Marras, F.~Kokkinos, J.~Deng, G.~Chrysos, and S.~Zafeiriou,
  ``Poly-nl: Linear complexity non-local layers with 3rd order polynomials,''
  in {\em Proc. IEEE/CVF Int. Conf. Comput. Vision}, pp.~10518--10528, October
  2021.

\bibitem{related:universal}
F.~Fan, J.~Xiong, and G.~Wang, ``Universal approximation with quadratic deep
  networks,'' {\em Neural Networks}, vol.~124, pp.~383--392, Jan. 2020.

\bibitem{related:deep_poly}
G.~G. Chrysos, S.~Moschoglou, G.~Bouritsas, J.~Deng, Y.~Panagakis, and
  S.~Zafeiriou, ``Deep polynomial neural networks,'' {\em IEEE Trans Pattern
  Anal Mach Intell}, vol.~44, no.~8, pp.~4021--4034, 2022.

\bibitem{theoretical:pruning}
L.~A. BRESLOW and D.~W. AHA, ``Simplifying decision trees: A survey,'' {\em
  Knowledge Eng Rev}, vol.~12, no.~01, p.~1–40, 1997.

\bibitem{2D_3D_conv}
K.~Bai, ``A comprehensive introduction to different types of convolutions in
  deep learning.''
  \href{https://towardsdatascience.com/a-comprehensive-introduction-to-different-types-of-convolutions-in-deep-}{https://towardsdatascience.com/a-comprehensive-introduction-to-different-types-of-convolutions-in-deep-learning-669281e58215}\\\href{learning-669281e58215}{https://towardsdatascience.com/a-comprehensive-introduction-to-different-types-of-convolutions-in-deep-learning-669281e58215},
  2019.
\newblock (accessed 11 April 2022).

\bibitem{intro:ablation}
R.~Meyes, M.~Lu, C.~W. de~Puiseau, and T.~Meisen, ``Ablation studies in
  artificial neural networks,'' {\em CoRR}, vol.~abs/1901.08644, 2019.

\bibitem{relu}
K.~Jarrett, K.~Kavukcuoglu, M.~Ranzato, and Y.~LeCun, ``What is the best
  multi-stage architecture for object recognition?,'' in {\em IEEE Int Conf
  Comput Vision}, pp.~2146--2153, 2009.

\bibitem{adam}
D.~P. Kingma and J.~Ba, ``Adam: A method for stochastic optimization,'' {\em
  CoRR}, vol.~abs/1412.6980, 2017.

\end{thebibliography}
\bibliographystyle{ieeetr}

\end{document}